\documentclass[review,12pt]{elsarticle}
\date{June 25, 2020}

\usepackage{amssymb}

\usepackage{multirow}
\usepackage[normalem]{ulem}
\useunder{\uline}{\ul}{}
\usepackage{subcaption} 
\usepackage{adjustbox}
\usepackage{fixfoot}
\usepackage{lscape}
\DeclareFixedFootnote\npirmnistfail{NPIR failed to scale to adapt to MNIST dataset even on a cloud server with 128 GB memory}
\DeclareFixedFootnote\ucidatasets{https://archive.ics.uci.edu/ml/index.php}
\DeclareFixedFootnote\elkidatasets{https://elki-project.github.io/datasets/}
\DeclareFixedFootnote\chameleondatasets{http://glaros.dtc.umn.edu/gkhome/cluto/cluto/download}
\DeclareFixedFootnote\mnistdatasets{http://yann.lecun.com/exdb/mnist/}
\DeclareFixedFootnote\keeldatasets{https://sci2s.ugr.es/keel/dataset.php?cod=183}
\DeclareFixedFootnote\benchmarkdatasets{ http://cs.joensuu.fi/sipu/datasets/}
\DeclareFixedFootnote\sklearn{https://scikit-learn.org/stable/}

\usepackage{longtable}
\usepackage{comment}
\usepackage[linesnumbered,ruled,vlined]{algorithm2e}
\usepackage{setspace}
\usepackage{float}
\usepackage{tabulary,xcolor}

\SetCommentSty{mycommfont}

\makeatletter
\newcommand*{\rom}[1]{\expandafter\@slowromancap\romannumeral #1@}
\makeatother

\usepackage{tabulary,xcolor}
\usepackage{amsfonts,amsmath,amssymb}
\usepackage[T1]{fontenc}
\usepackage{comment}

\journal{Pattern Recognition}

\begin{document}

\begin{frontmatter}
	
\title {DenMune: Density Peak Based Clustering Using Mutual Nearest Neighbors}              
    
\author[a8750c9237b3b]{Mohamed Abbas}
\ead{mohamed.alyabbas@alexu.edu.eg}
\author[a8750c9237b3b]{Adel El-Zoghabi}
\ead{adel.elzoghby@alexu.edu.eg}
\author[ae5f2991a77a3,adde06d18deeb]{Amin Shoukry}
\ead{amin.shoukry@ejust.edu.eg}
    
\address[a8750c9237b3b]{Information Technology\unskip, 
    Institute of Graduate Studies and Research\unskip, Alexandria\unskip, Egypt}
  	
\address[ae5f2991a77a3]{Computer Science and Engineering\unskip, 
    Egypt-Japan University of Science and Technology\unskip, New Borg-El-Arab City\unskip, Alexandria\unskip, Egypt}

\address[adde06d18deeb]{Computer and Systems Engineering Dept\unskip, 
    Faculty of Engineering\unskip, Alexandria\unskip, Egypt}

\begin{abstract}
Many clustering algorithms fail when clusters are of arbitrary shapes, of varying densities, or the data classes are unbalanced and close to each other, even in two dimensions. A novel clustering algorithm "DenMune" is presented to meet this challenge. It is based on identifying dense regions using mutual nearest neighborhoods of size $K$, where $K$ is the only parameter required from the user, besides obeying the mutual nearest neighbor consistency principle. The algorithm is stable for a wide range of values of $K$. Moreover, it is able to automatically detect and remove noise from the clustering process as well as detecting the target clusters. It produces robust results on various low and high dimensional datasets relative to several known state of the art clustering algorithms. 
\end{abstract}

\end{frontmatter}

\section{Introduction}
Data clustering, which is the process of gathering similar data samples into groups/clusters, has been found useful in different fields such as medical imaging (to differentiate between different types of tissues \unskip~\cite{medical_applications_2018}), market research (to partition consumers into perceptual market segments \unskip~\cite{customers_segmentation_2018}), document retrieval (to find documents that are relevant to a user query in a collection of documents  \unskip~\cite{document_retrieval_2018}), and fraud detection (to detect suspicious fraudulent patterns) \unskip~\cite{fraud_detection_2019}), as well as many others  \unskip~\cite{clustering_survey_2013}. In general, Clustering algorithms can be divided into the following types:

\subsection{Partitioning-based Clustering Algorithms}
\label{subsec:Outline1}
In this category, data objects are divided into  non-overlapping subsets (clusters) such that each object lies in exactly one subset. The most well-known and commonly used algorithm in this class is K-means. K-means is heavily dependent on the initial cluster centers, which are badly affected by noise and outliers. A well known variant is K-medoid. K-medoid selects  the most centrally located point in a cluster, namely its medoid, as its representative point. Another well-known variant of K-means is KMeans++. It chooses centers at random, but weighs them according to the square distance from the closest already chosen center.

A recent algorithm in this area is RS algorithm \unskip~\cite{rs_2018}. It belongs to the class of  swap-based clustering algorithms that aim at using a sequence of prototype swaps to deal with the inability of K-means in fine-tuning the cluster boundaries globally, although it succeeds locally. By adopting a random swap strategy the computational complexity is reduced  and the results are better than those obtained by k-means. Its main limitation is that there is no clear rule how long the algorithm should be iterated.

Another recent algorithm in this category is CBKM \unskip~\cite{cbkm_2019}. It investigates the extent to which using better initialization (poor initialization can cause the algorithm to get suck at an inferior local minimum) and repeats can improve the k-means algorithm. It is found that when the clusters overlap, furthest point heuristic(Maxmin)can reduce the number of erroneous clusters from 15

\subsection{Proximity-based Clustering Algorithms}
\label{subsec:Outline4}
Neighborhood construction is useful in  discovering the hidden interrelations between connected patterns  \unskip~\cite{apollonius_2018}.  Proximity can be identified using k-nearest-neighbor (cardinality-based), or identified using $\epsilon$ -neighbourhood (distance-based).

A recent algorithm in this category is FastDP algorithm \unskip~\cite{fastdp_2019}. It focuses on improving the quadratic time complexity of the "Density peaks" popular clustering algorithm by using a fast and generic construction of approximate k-nearest neighbor graph both for density and for delta calculation (distance to the nearest point with higher density).
The cluster centers are selected so that they have a high value of both delta and density. After that, the remaining points are allocated (joined) to the already formed clusters by merging with the nearest higher density point. The algorithm inherits the problems associated with the original "Density peaks" algorithm which include: (1) how to select the initial k cluster centers based on dendity and delta. The algorithm adopts the gamma strategy (which uses the points with high product of the two features(density and delta). (2) the problem of how to threshold the density and delta features.

 Another recent algorithm is NPIR algorithm \unskip~\cite{npir_2020}. it finds the nearest neighbors for the points that are already clustered based on the Euclidean distance between them and cluster them accordingly.  Different nearest neighbors are selected; from the kNN lists of the already clustered point; at different iterations of the algorithm. Therefore,the algorithm relies on the random and iterative behavior of the partitional clustering algorithms to give quality clustering results. It performs Election, Selection, and Assignment operations to assign data points to appropriate clusters. Therefore, three parameters should are needed: The number of clusters, The indexing ratio (controls the amount of possible reassignment of points) and the number of iterations.

CMUNE \unskip~\cite{cmune}, a predecessor of DPC, uses the MNN  graph to calculate the density of each point and select the high-density points (also called strong points) as the seeds from which clusters may grow up. A cutoff parameter is also needed to differentiate between strong and  weak points. Similar to DPC, the constructed clusters are very sensitive to variations in this parameter. The notion of weak/isolated points has been introduced in (\unskip~\cite{cmune}, \unskip~\cite{csharp} and \unskip~\cite{CHKNN-2018}) to define points which are prone to be classified as noise and, consequently, excluded from the clusters' formation. 

\subsection{Hierarchical Clustering Algorithms}
\label{subsec:Outline2}
In this category, data objects are organized into a tree of group-of-objects. The tree is constructed either from top to bottom or from bottom to top leading to divisive or agglomerative type of algorithms, respectively. Hierarchical clustering has been extensively applied in pattern recognition. Some known examples are Chameleon\unskip~\cite{chameleon} and CURE \unskip~\cite{cure}. The scalability of hierarchical methods is generally limited due to their time complexity. To address this issue, \unskip~\cite{fast_HC_2018} proposed a fast hierarchical clustering algorithm based on topology training. 
PHA \unskip~\cite{pha_2013} uses both local and global data distribution information during the clustering process. It can deal with overlapping clusters, clusters of non-spherical shapes and clusters containing noisy data, by making good use of the similarity between the iso-potential contours of a potential field and  hierarchical clustering.
A more successful variant of hierarchical density is HDBSCAN \unskip~\cite{hdbscan_2017}. HDBSCAN provides a clustering hierarchy from which a simplified tree of significant clusters is constructed, then a flat partition composed of clusters extracted from optimal local cuts through the cluster tree. Unlike DBSCAN, It can find clusters of variable densities.\\

RCC \unskip~\cite{rcc_2017} is a clustering algorithm that achieves high accuracy across multiple domains and scales efficiently to high dimensions and large datasets. it optimizes a smooth continuous objective function that allows the algorithm to be extended to perform joint clustering and dimensionality reduction.

A recent algorithm in this category is FINCH algorithm \unskip~\cite{finch_2019}. It is fully parameter-free (i.e. does not require any user defined parameters such as similarity thresholds, number of clusters or a priori knowledge about the data distribution) clustering algorithm.The algorithm is based on the clustering equation which defines an adjacency link matrix that links two points i and j if j is the first neighbor of i or i is the first neighbor of j or both i and j have (share) the same first nearest neighbor. The algorithm belongs to the family of hierarchical agglomerative methods, has low computational overhead and is fast.

In this paper, a novel clustering algorithm DenMune is presented for the purpose of finding complex clusters of arbitrary shapes and densities in a two-dimensional space. Higher dimensional spaces are first reduced to 2-D using the t-sne algorithm. It can be considered as a variation of the CMUNE algorithm \unskip~\cite{cmune}. DenMune requires only one parameter from the user across its two-phases. Other advantages include its ability in automatically detecting, removing and excluding noise from the clustering process. It adopts a voting-system where all data points are voters but only those that receive highest votes are considered clusters' constructors. Moreover, it automatically detects the target clusters and  produces robust results with no cutoff parameter needed.

\subsection{Outline of the Paper}
\label{subsec:Outline}

The rest of this paper is organized as follows. Section 2 describes the key concepts of the DenMune clustering algorithm. Section 3 describes the algorithm itself and its time complexity analysis. Section 4 presents the data sets used in the experiments conducted to evaluate the performance of the algorithm. Section 5 presents the conclusion and possible future work.

\section{Basic Definitions and Mechanisms Underlying the Proposed Algorithm}
\label{sec:BASIC_DEFINITIONS}

In this section we describe the basic concepts used in the proposed algorithm and its underlying mechanisms.

\subsection{K-Mutual-Neighbors Consistency}
\label{subsec:K-MNB}

The principle of K-Mutual-Neighbors (K-MNN) consistency \unskip~\cite{disconnectivity}; which states that for any data points in a cluster its \textit{MNN} should also be in the same cluster; is stronger than the K-nearest Neighbors (KNN) consistency concept. In CMune and CSharp (\unskip~\cite{cmune},  \unskip~\cite{csharp}) the concept of K-MNN is used to develop a clustering framework based on "Reference Points", defined in section  \ref{subsec:Reference-List_DEFINITION}, in which dense regions are identified using mutual nearest neighborhoods of size $K$, where $K$ is a user-parameter. Next, sets of points sharing common mutual nearest neighborhoods are considered in an agglomerative process to form the final clusters. This process is controlled by two threshold parameters. In contrast, by properly partitioning the data points into classes (section \ref{subsec:Type_of_points}) and guided by the principle of K-Mutual-Neighbors consistency (K-MNN), DenMune is able to get rid of these threshold parameters in performing its clustering task (section \ref{subsec:Conservative_nature_of_DenMune}).

\subsection{Refer-To-List, Reference-List and Reference Point}
\label{subsec:Reference-List_DEFINITION}

Given a set of points $P = $ \{$p_1, p_2,$ \ldots  $p_{n-2}, p_{n-1}, p_n$ \}, let $KNN_{p_i \rightarrow} = \{p_1, p_2 ,p_3,\ldots, p_k \}$ be the K-nearest neighbors of point $p_i$. In this paper, we consider that  points in a $KNN$ set are sorted, ascendingly, according to their distances from a given  reference point. Therefore,  $KNN_{p_i \rightarrow}$ represents the ordered list of points that $p_i$ refers-to, namely, the  "Refer-To List".
If \begin{math}p_i \in \end{math} $KNN_{p_j \rightarrow}$, then $p_i$ is referred-to by $p_j$. In this case,  \begin{math}p_j \in \end{math} $KNN_{p_i \leftarrow}$,  the set of points considering $p_i$ among their K-nearest neighbors.   The set $KNN_{p_i \rightarrow} \cap KNN_{p_i \leftarrow}$, is the set $MNN_{p_i}$ of mutual nearest neighbors of ${p_i}$. It represents a set of dense points associated with point $p_i$. Point $p_i$ is said to be the "Representative-Point"  or  "Reference-Point" of $MNN_{p_i}$. 

\begin{figure}[!ht]
\centering
\includegraphics[height =6.0 cm]{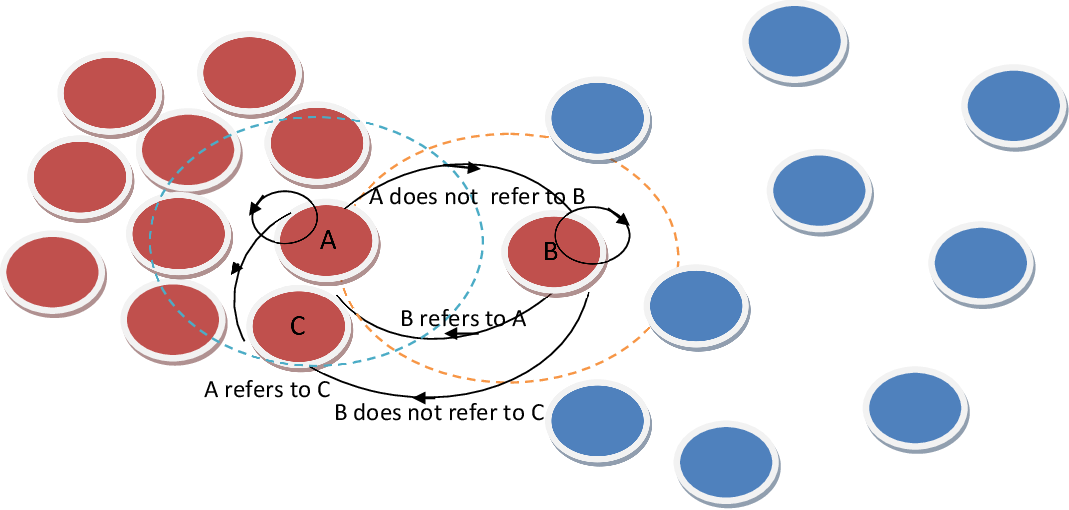}
\caption {Asymmetry of the \textit{K}-nearest neighborhood relation.}
\label{main-idea}     
 \end{figure}

As shown in  Fig. \ref {main-idea}, although the Euclidean
distance is a symmetric metric, from the SNN \unskip~\cite{snn}
perspective (and considering $K=4$), point $A$ is in $KNN_{B_\rightarrow}$, however, $B$ is
not in $KNN_{A_\rightarrow}$.

\subsection{DenMune classification of data points into Strong, Weak and Noise Points}
\label{subsec:Type_of_points}
According to the value of the  non-negative ratio \(r=\frac{|KNN_{p\leftarrow}|}{|KNN_{p\rightarrow}|}= \frac{|KNN_{p\leftarrow}|}{K} \), since 
\begin{math} |KNN_{p\rightarrow}| = K \end{math} (by definition), from DenMune point of view, each data point 'p' in a  dataset, belongs to one of the types described in Eq.(\ref{EQ:Types_of_Points}):

\begin{equation} p.Type =
  \begin{cases}
  \text {Strong point}  & \text{if  $r \geq 1$} \\
 \text {Weak point}   & \text{if  $r < 1$} \\
    \text  {Noise point} & \text{if  $ 0 \leq  r \ll 1$}
  \end{cases}
  \label{EQ:Types_of_Points}
 \end{equation}

\begin{figure}[!ht]
\centering
\includegraphics[height=5.0cm]{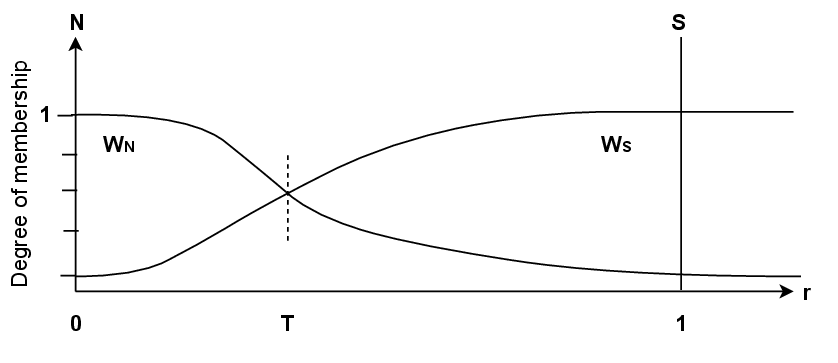}
\caption {Fuzziness of the  set $W$ of weak points. $N$ and $S$ denote the noise ($r = 0)$ and strong ($r \geq 1$) points, respectively. $T$ is some threshold that partitions the set $W$ into $W_N$ and $W_S$. Both sets are automatically detected by DenMune.}
\label{Fig:fuzzy-set}      
 \end{figure}
 
\begin{itemize}
\item Strong Points: satisfy the condition
 \begin{math} |KNN_{p\leftarrow}| \geq |KNN_{p\rightarrow}|  \end{math}, or
 \begin{math} |KNN_{p\leftarrow}| \geq K  \end{math}.  This implies that $|MNN_{p}| =$| $KNN_{p\rightarrow}$ $\cap $ $KNN_{p\leftarrow}|$ $=$ \textit{K}.
Strong points are also called seed points. Seed points that share non-empty \textit{MNN}-sets of seeds are the clusters' constructors in the proposed algorithm.

\item Weak points: satisfy the condition
 \begin{math} |KNN_{p\leftarrow}| < |KNN_{p\rightarrow}|  \end{math}. From Eq.(\ref{EQ:Types_of_Points}), it is clear that the boundaries of the set defining the weak points are fuzzy. Fig. \ref{Fig:fuzzy-set} illustrates the idea that in DenMune, a weak point either succeeds in joining a cluster or it is considered as noise. For this reason,  weak-points are called  non-strong (non-seed) points.  Hence, the following lemma can be concluded:

\underline {Lemma}: The set of weak points is a fuzzy set. Its boundaries with the sets of strong and noise points are fuzzy. The rule governing the assignment of a weak point to a cluster or rejecting it as noise is, in general, data as well as algorithm dependent.

\item Noise points, represent points either with empty $MNN$s (corresponding to $r= 0$, which are removed early in phase \rom{1} of DenMune algorithm, named as noise of type-1), or weak points that fail to merge with any formed cluster (corresponding to $r \ll 1 $, which are removed in phase \rom{2} of the algorithm, named as noise of type-2).
\end{itemize}

\subsection{Proposed Algorithm: Overview}
\begin{figure}[!ht]
\begin{minipage}[b]{\linewidth}
%
\begin{subfigure}[t]{0.49\textwidth}
       \raisebox{-\height}{\includegraphics[width=\textwidth, height=2.8 cm]{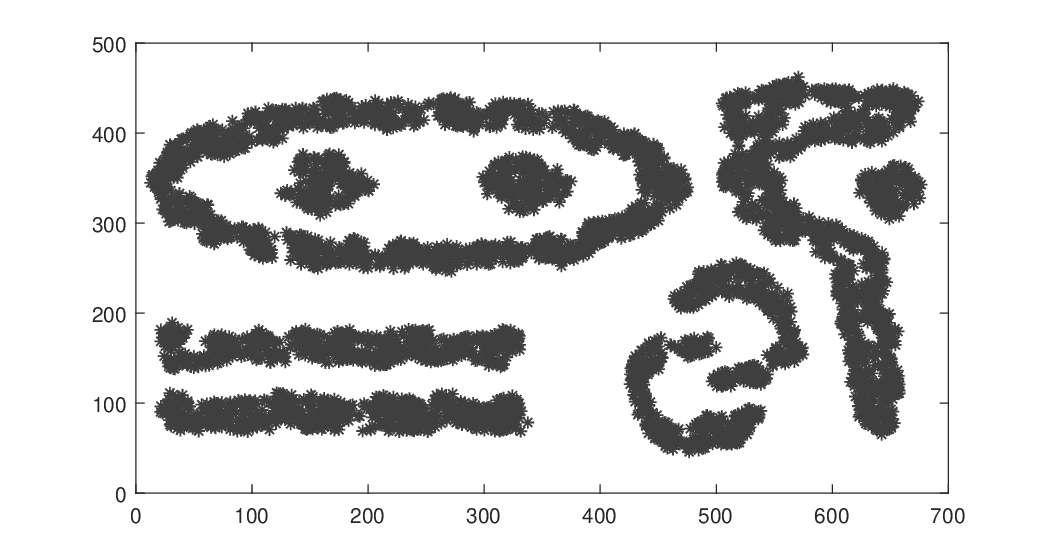}}
        \caption{Backbone-constructors (points that receive high votes), also known as strong points. }
        \label{strong-points}
\end{subfigure}
  \hfill
\begin{subfigure}[t]{0.49\textwidth}
        \raisebox{-\height}{\includegraphics[width=\textwidth, height=2.8 cm]{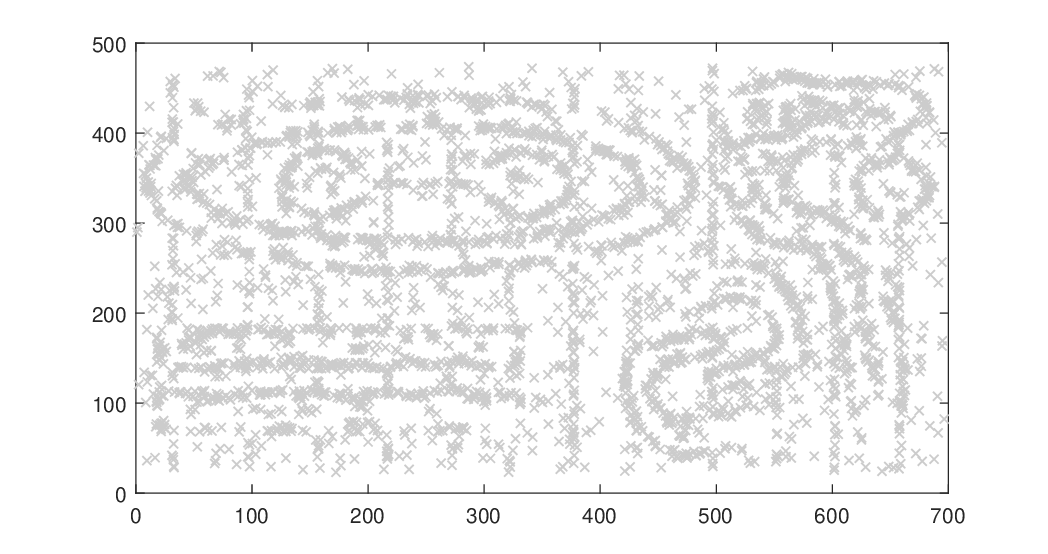}}
        \caption{Weak points.}
         \label{weak-points}
\end{subfigure}
\hfill
\begin{subfigure}[t]{0.49\textwidth}
	     \raisebox{-\height}{\includegraphics[width=\textwidth, height=2.8 cm]{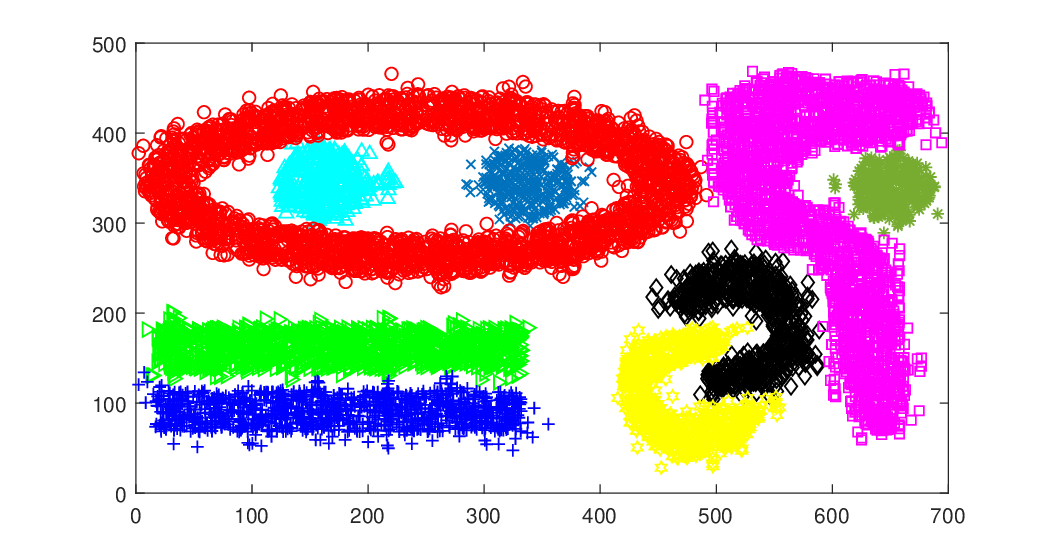}}
        \caption{DenMune merges some of the weak points in Fig. \ref{weak-points} with their nearest clusters.}
\end{subfigure}
\hfill
\begin{subfigure}[t]{0.49\textwidth}
       \raisebox{-\height}{\includegraphics[width=\textwidth, height=2.8 cm]{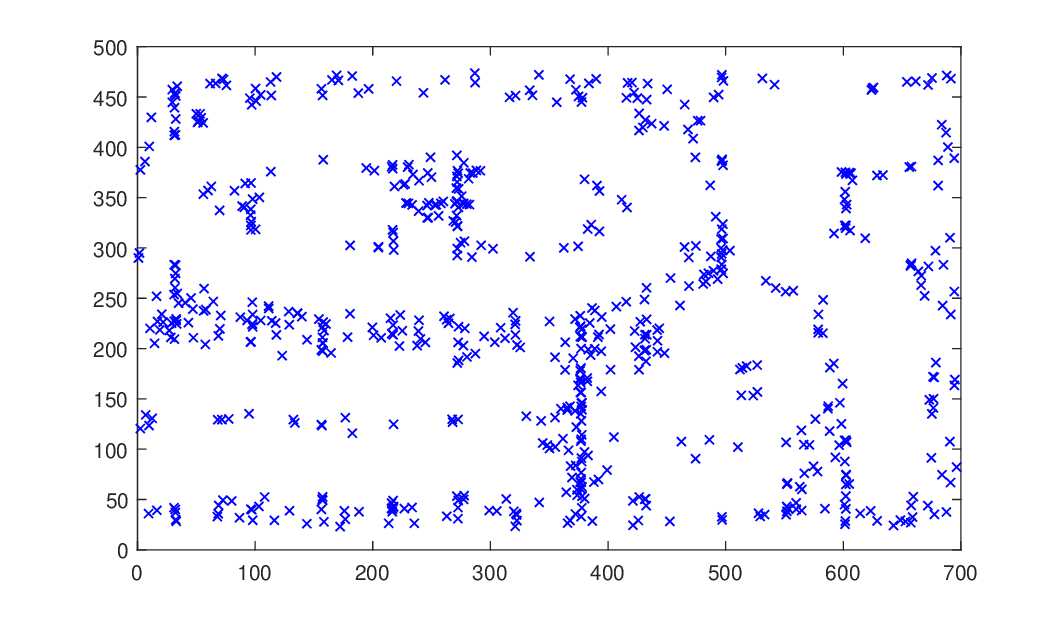}}
        \caption{Noise points.}
\end{subfigure}
\hfill
\end{minipage}%
\caption{Phases of DenMune }
\label{Fig:DenMune-Merging-Process}
\end{figure}

DenMune is based on a voting system framework where points that receive the largest number of votes (i.e. they belong to the K-nearest neighbors of at least \textit {K} other points), are marked as dense/ seed points and are used to construct the backbone of the target clusters in phase \rom{1} of the algorithm. Points that receive no votes are considered as noise of type-1 and are eliminated from the clustering process. Phase \rom{2} deals with the weak points that either survive by merging with the existing clusters, or are eliminated by being considered as noise of type-2.
\\
Table \ref{tab:points-spread} shows the distribution of strong/ seeds and weak/ non-seeds points among the  Chameleon's DS7 dataset which includes 10,000 data points, while Fig. \ref{Fig:DenMune-Merging-Process} illustrates how strong points determine the shapes/ structures of the clusters where weak points can only merge with them. 

\begin{table}[!ht]
 \centering
    
    \caption{Strong and weak points found by DenMune in the Chameleon DS7 dataset.}
    \label{tab:points-spread}
\begin{tabular}{cccccc}%
  
   \hline
    \hline
   Algorithm & Parameters &  Strong Points & Weak Points & Noise of type-1 & noise of type-2\\
    \hline 
   DenMune & $K$=39 & 5858 & 3471 & 0 & 671 \\
\hline

\end{tabular}
  \end{table}

\subsection{Proposed Algorithm: Steps}
DenMune involves the following steps:
\begin{itemize}
\item Canonical ordering: Clustering results obtained by DenMune are deterministic, as it orders the set of points $P$ according to $|KNN_{p\leftarrow}|$ in a descending order.

\item Noise Removal: Noise points of type-1 as well as those of type-2 are detected and removed in phase \rom{1}   and phase \rom{2}, of the algorithm, respectively, as illustrated in Table \ref{tab:noises-types}.

\item Skeleton Construction and Propagation: after  removal of type-1 noise points, the remaining points are partitioned into two groups: dense points (seeds) and low-dense points (non-seeds). Only seed points are eligible to construct the skeleton of the target clusters (i.e. the number of seed points represent an upper bound on the number of clusters), while low-dense points are considered in the next phase.

\begingroup
\setlength{\tabcolsep}{8pt} 
\renewcommand{\arraystretch}{0.73} 
\centering

\begin{table}[H]
\caption{Distribution of the different type of points, detected by DenMune, vs the number \textit{K} of nearest neighbors in Chameleon DS7 dataset.}
\label{tab:noises-types}

\centering
\begin{tabular}{ccccc}
 \hline
 $K$ & Strong Points &
  Weak Points &  Noise of type-1 &
Noise of type-2 \\ 
\hline
 \hline
1  & 6078 & 0    & 3922 & 0    \\
2  & 6545 & 958  & 1200 & 1297 \\
3  & 6448 & 1910 & 369  & 1273 \\
4  & 6262 & 2572 & 135  & 1031 \\
5  & 6110 & 2933 & 71   & 886  \\
6  & 6013 & 3164 & 45   & 778  \\
7  & 5968 & 3334 & 36   & 662  \\
8  & 5955 & 3400 & 28   & 617  \\
9  & 5896 & 3485 & 17   & 602  \\
10 & 5866 & 3589 & 13   & 532  \\
11 & 5899 & 3572 & 10   & 519  \\
12 & 5826 & 3668 & 5    & 501  \\
13 & 5830 & 3635 & 5    & 530  \\
14 & 5820 & 3643 & 4    & 533  \\
15 & 5819 & 3638 & 4    & 539  \\
16 & 5809 & 3572 & 4    & 615  \\
17 & 5833 & 3550 & 4    & 613  \\
18 & 5854 & 3539 & 4    & 603  \\
19 & 5829 & 3556 & 4    & 611  \\
20 & 5814 & 3568 & 3    & 615  \\
\end{tabular}
\end{table}
\endgroup

To further illustrate the process of clusters propagation, Chameleon's dataset DS7 \chameleondatasets{}{} is
used. Several snapshots of the clustering process, are shown in Fig. \ref{Fig:DenMune-Propagation}, to illustrate how clusters propagate agglomeratively, and in parallel, in CSharp and DenMune.

\begin{figure}[H]

\begin{minipage}[b]{\linewidth}
\centering

\begin{subfigure}[ht]{0.49\textwidth}
        \raisebox{-\height}{\includegraphics[width=\textwidth, height=2.9 cm]{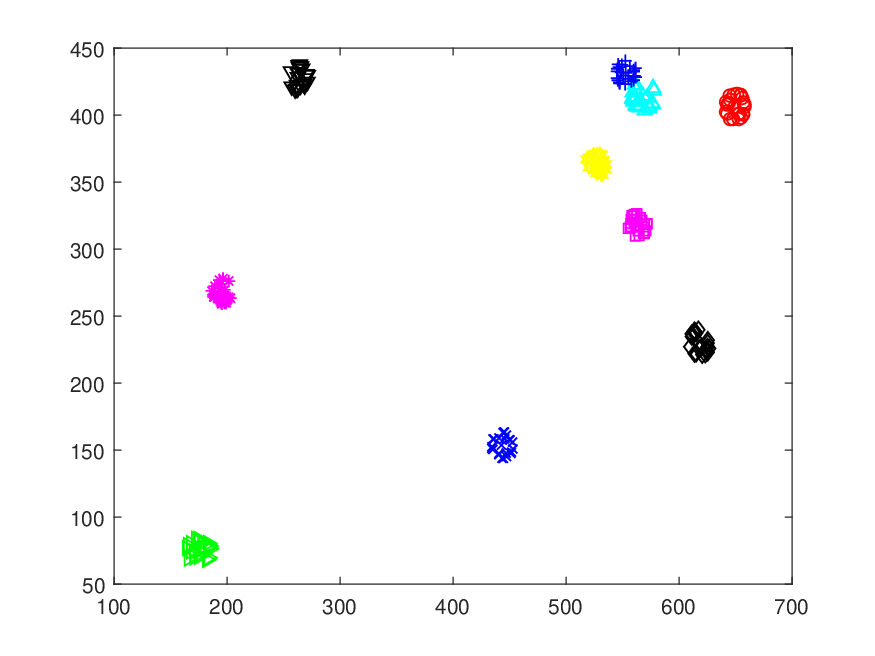}}
        \caption{CSharp: at the $ 10^{th}$ iteration}
\end{subfigure}
  \hfill
\begin{subfigure}[ht]{0.49\textwidth}
        \raisebox{-\height}{\includegraphics[width=\textwidth,  height=2.9 cm]{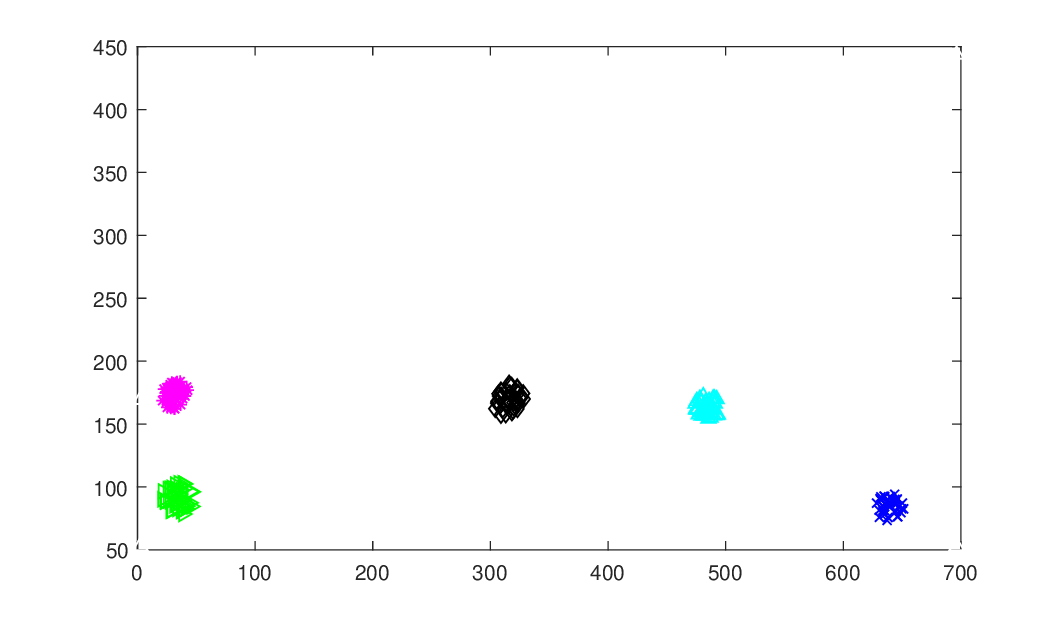}}
        \caption{DenMune : at the $ 10^{th}$ iteration}
\end{subfigure}
\hfill

\begin{subfigure}[ht]{0.49\textwidth}
        \raisebox{-\height}{\includegraphics[width=\textwidth, height=2.9 cm]{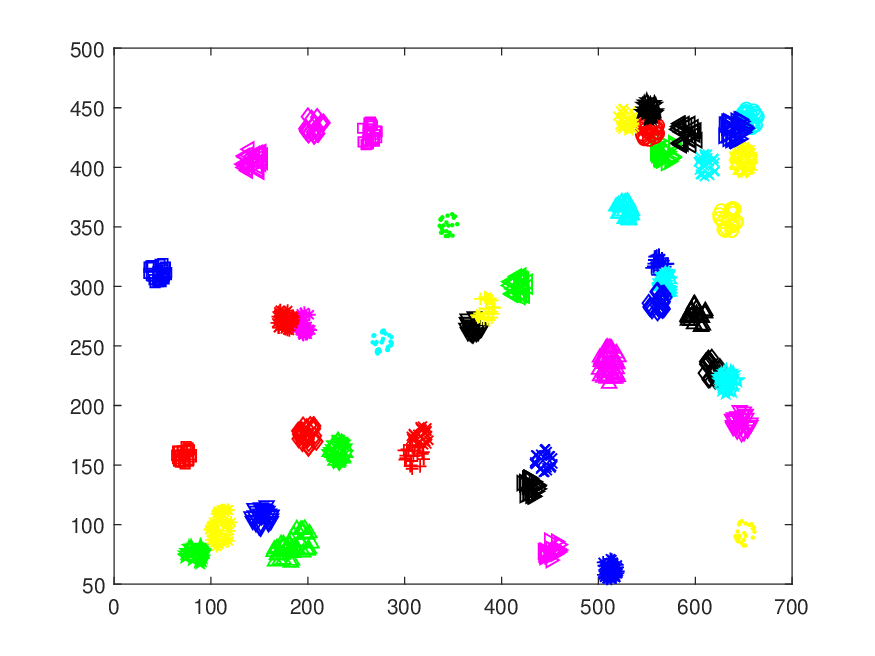}}
        \caption{CSharp: at the $ 50^{th}$ iteration}
\end{subfigure}
  \hfill
\begin{subfigure}[ht]{0.49\textwidth}
       \raisebox{-\height}{\includegraphics[width=\textwidth,  height=2.9 cm]{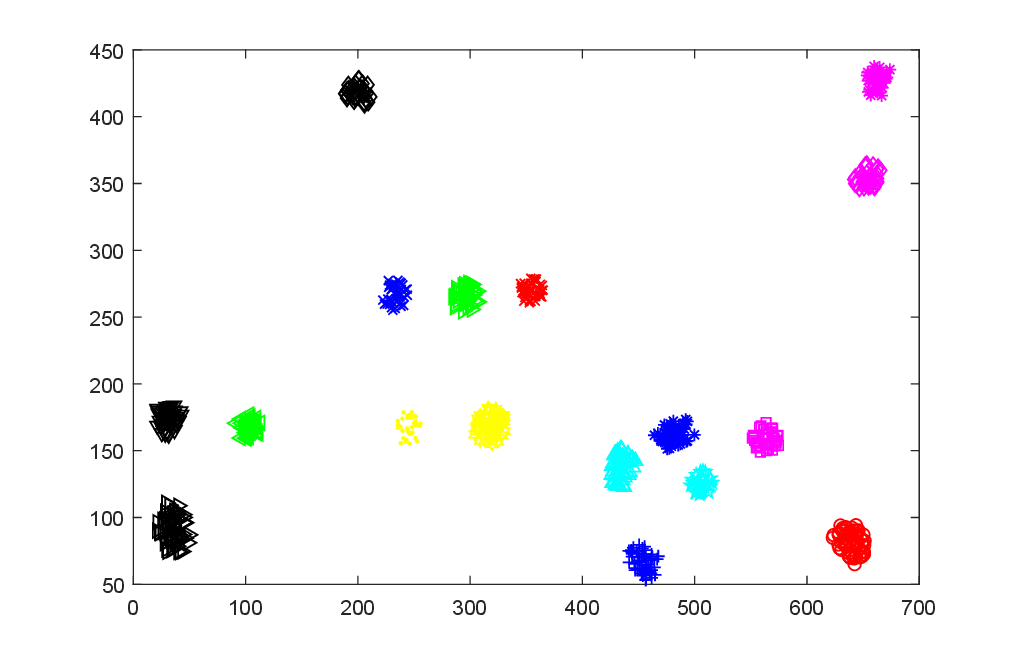}}
       \caption{DenMune : at the $ 50^{th}$ iteration}
\end{subfigure}
\hfill
\begin{subfigure}[ht]{0.49\textwidth}
        \raisebox{-\height}{\includegraphics[width=\textwidth,  height=2.9 cm]{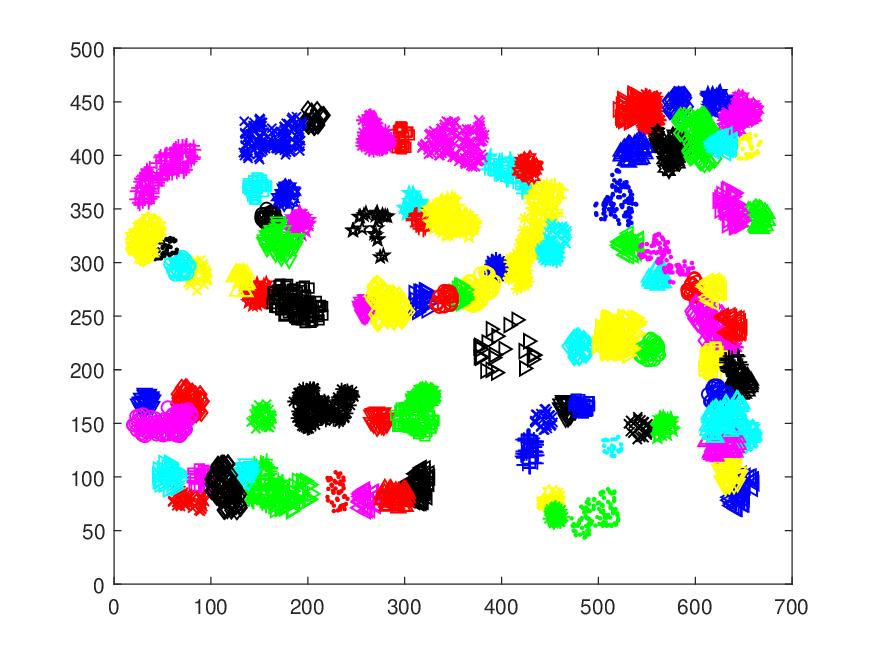}}
       \caption{CSharp: at the $ 250^{th}$ iteration}
\end{subfigure}
    \hfill
\begin{subfigure}[ht]{0.49\textwidth}
        \raisebox{-\height}{\includegraphics[width=\textwidth,  height=2.9 cm]{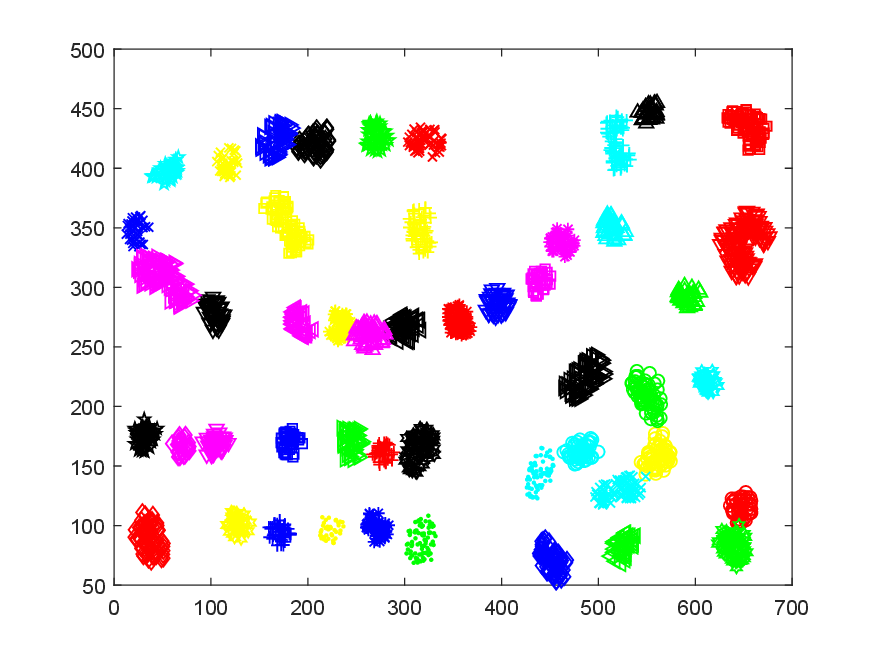}}
        \caption{DenMune : at the $ 250^{th}$ iteration}
\end{subfigure}
\hfill

\begin{subfigure}[ht]{0.49\textwidth}
        \raisebox{-\height}{\includegraphics[width=\textwidth,  height=2.9 cm]{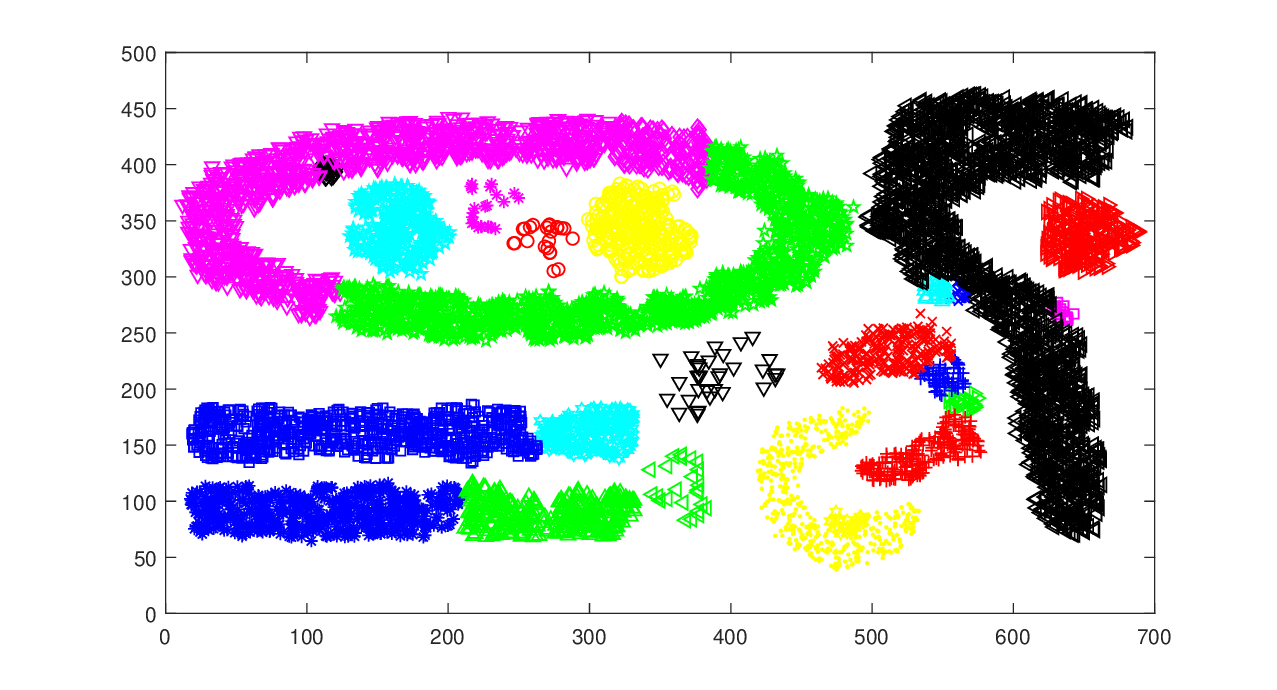}}
         \caption{CSharp: at the $ 1000^{th}$ iteration}
\end{subfigure}
    \hfill
\begin{subfigure}[ht]{0.49\textwidth}
        \raisebox{-\height}{\includegraphics[width=\textwidth,  height=2.9 cm]{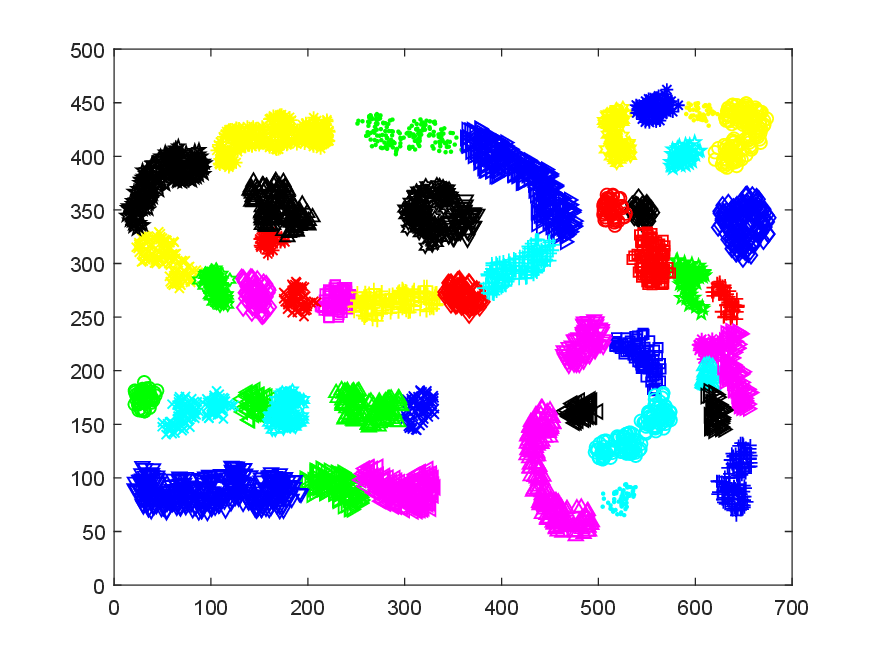}}
        \caption{DenMune : at the $ 1000^{th}$ iteration}
\end{subfigure}
\hfill

\begin{subfigure}[ht]{0.49\textwidth}
        \raisebox{-\height}{\includegraphics[width=\textwidth,  height=2.9 cm]{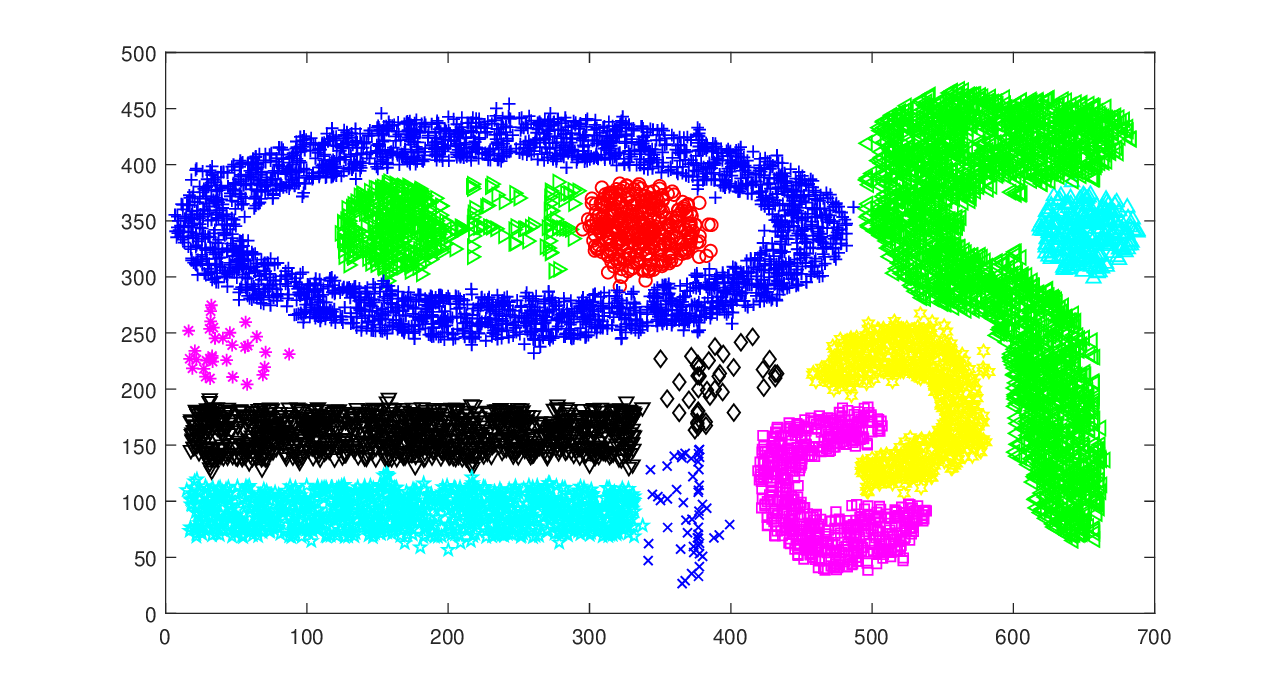}}
        \caption{CSharp: at the last iteration, $6734^{th}$}
\end{subfigure}
    \hfill
\begin{subfigure}[ht]{0.49\textwidth}
        \raisebox{-\height}{\includegraphics[width=\textwidth,  height=2.9 cm]{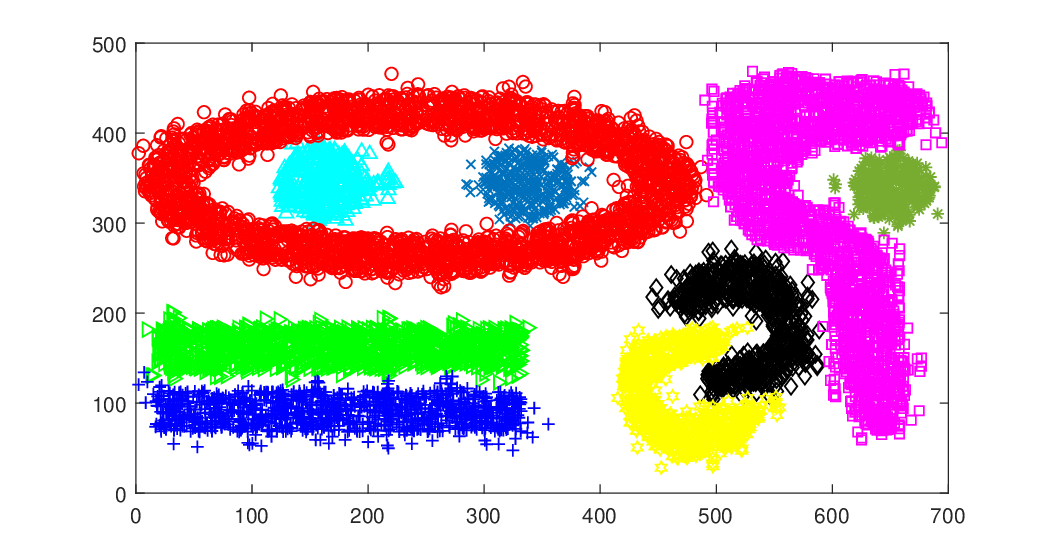}}
        \caption{DenMune: at last iteration, $9329^{th}$}
\end{subfigure}
\hfill

\caption{Clusters formation and propagation in DenMune and CSharp. Clusters seeds in DenMune are sparser but their propagation speed is slower. Also, DenMune results are more noise free. }
\label{Fig:DenMune-Propagation}
\end{minipage}
\end{figure}

\subsection{Conservative Nature of DenMune}
\label{subsec:Conservative_nature_of_DenMune}

\begin{itemize}
\item Clusters formation in Phase \rom{1}: Fig. \ref{Fig:clusters_Num_Prop}(a), illustrates the evolution of the number of clusters with the number of iterations for Chameleon's dataset. DenMune merges clusters conservatively in contrast to CSharp which is eager to merge clusters. Table \ref{tab:points-spread}, indicates that 5858 strong points are found by DenMune during this phase.  Therefore, the process of clusters formation stabilizes after $5858$  iterations at the end of phase \rom{1}, after which no more clusters can be constructed.

\item Slow Merging of Weak Points in Phase \rom{2}: weak  points are merged one by one, each to the cluster with which it shares the largest number of $MNN$-seeds. Table \ref{tab:points-spread}, indicates that out of the $4142 $  $(3471 + 671)$ weak points, 3471 of them succeed in merging with the clusters formed in the first phase. The remaining 671 points are considered as noise points of type-2. It is worth to note that  DenMune overcomes the lack of the noise threshold $L$ and the merge parameter $M$, used in CSharp, by (1) strengthening the $MNN$ relationship to involve only seed points, (2) the propagation process considers the weak  points individually, i.e. one by one, (3) weak points that fail to merge with the formed clusters are detected and removed as noise. As shown in Fig. \ref{Fig:clusters_Num_Prop}(b), for the DS7 dataset, after 1000 iterations, CSharp clustered 80\% of the data points, while DenMune clustered only 50\% of them. This is due to the fact that clusters in DenMune are initially sparse, as shown in Fig. \ref{Fig:clusters_Num_Prop}(b).
\end{itemize}

\end{itemize}

\begin{figure}[H]
\begin{minipage}[b]{\linewidth}
%
\begin{subfigure}[t]{0.45\textwidth}
 	\raisebox{-\height}{\includegraphics[width=\textwidth, height=4.5 cm]{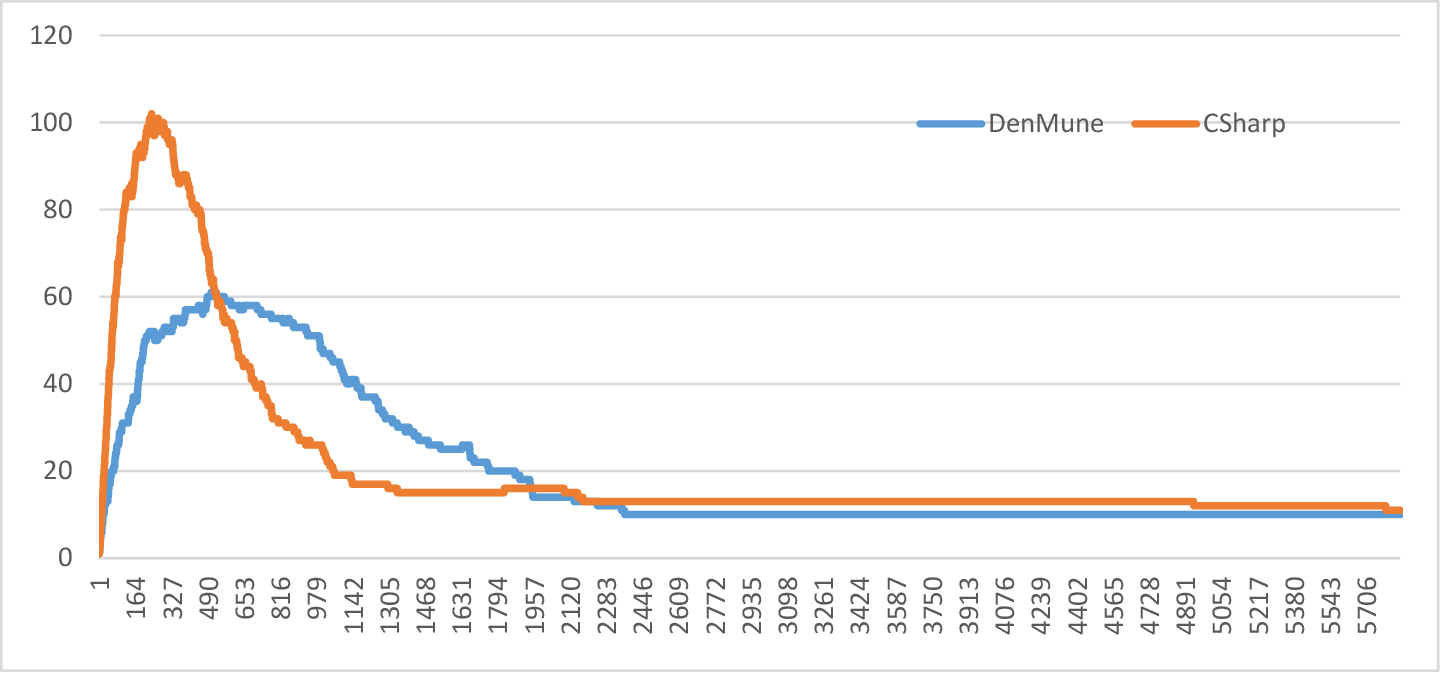}}
     \caption{Number of clusters vs number of iterations.}
       
\end{subfigure}
  \hfill
\begin{subfigure}[t]{0.45\textwidth}
	\raisebox{-\height}{\includegraphics[width=\textwidth, height=4.5 cm]{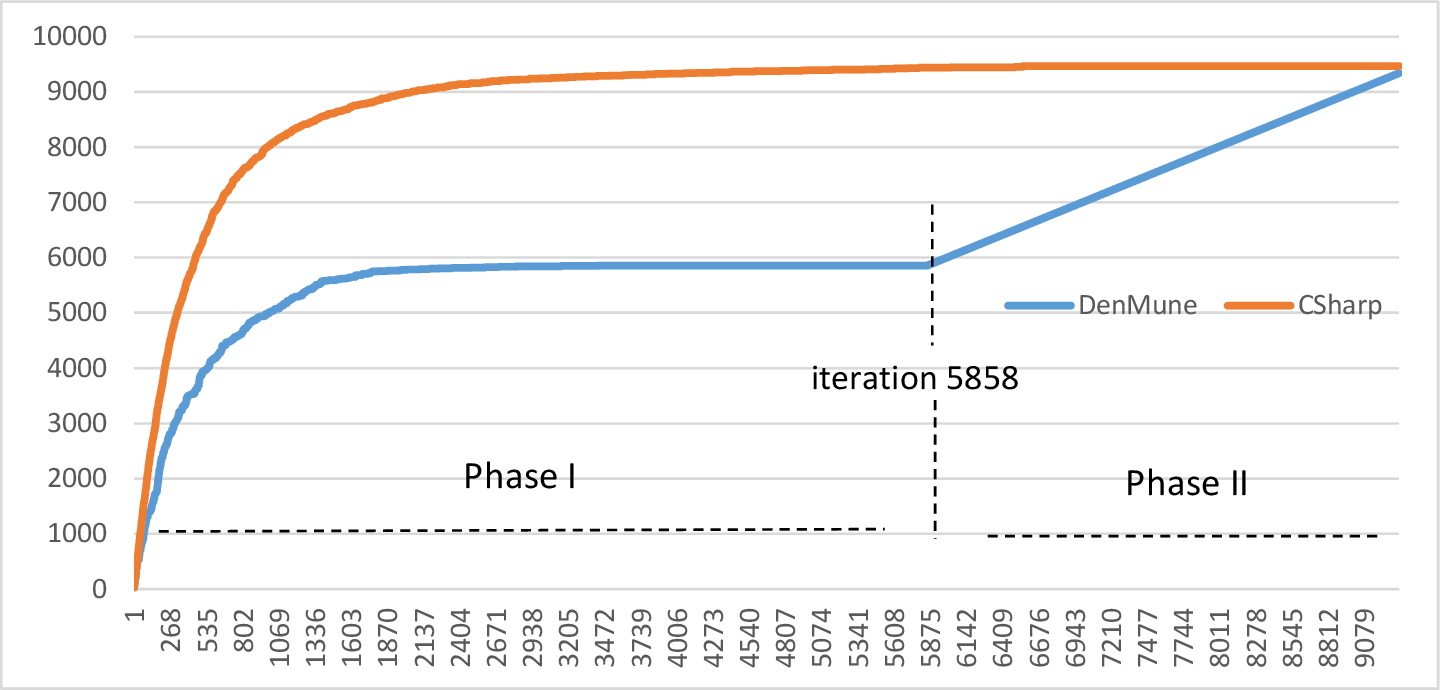}}
     \caption{Number of clustered data points vs number of iterations.}
    \end{subfigure}
\hfill
\end{minipage}%
\caption{DenMune vs CSharp:(a) Number of clusters and (b) number of clustered data points vs number of iterations. }
\label{Fig:clusters_Num_Prop}
\end{figure}%

\section{DenMune Algorithm}
\label{section:DenMune_Algorithm}

Algorithm \ref{Alg:DenMune} describes the proposed algorithm, followed by a
detailed discussion of its time complexity. 

\begin{algorithm}[H]
\DontPrintSemicolon 
\KwIn{ Data points $P = \{p_1, p_2\ldots,p_n \}$, $K$ \tcp {size of the neighborhood of a point}}
\KwOut { $C$ \tcp{ set of generated clusters }}
Construct distance matrix $D$\;
\tcp {Construct the Refer-To-List, $KNN_{p_i\rightarrow}$, for each point $p_i \in P$}
$KNN_{p_i\rightarrow} \gets \{j | d(p_i,p_j) \le d(p_i, p_k) \}$\;
\tcp{For each point $p_i$  construct $KNN_{p_i\leftarrow}$ by scanning $KNN_{p_j\rightarrow}$ and selecting points $j$ having point $i$ in their $KNN_{p_j\rightarrow}$}

\ForEach{ $p_i \in P$ }{
    \ForEach{ $p_j \in P$ }{
     
     \If{$p_i \in  KNN_{p_j\rightarrow}$}{
       $KNN_{p_i\leftarrow} \gets \{ KNN_{p_i\leftarrow}$   $\cup$ ${p_j}$ \}\;
        }
     }

\tcp{From $KNN_{p_i\rightarrow}$  and  $KNN_{p_i\leftarrow}$, construct $MNN_{p_i}$ }
$MNN_{p_i} \gets KNN_{p_i\rightarrow}$ $\cap$  $KNN_{p_i\leftarrow}$ 

}

Remove the set $O$, of noise points $p_i$ of type-1, satisfying $|MNN_{p_i}| = 0$ \;
Form the sorted  list $P$, The sorting is in a descending order according to $|KNN_{p_i\leftarrow}|$ \tcp{$P$ = $P$ - $O$ }
Form the sorted  list $S \subset P$ $=$ \{$p_i | p_i$ satisfies $|KNN_{p_i\leftarrow}|$ $\geq$  $|KNN_{p_i\rightarrow}|\}$\; 

Form the set $Q$ of non-seed points, where $Q = P - S$ \tcp{Note that $Q \subset P$ $=$ \{$p_i | p_i$ satisfies $|KNN_{p_i\leftarrow}| <$   $|KNN_{p_i\rightarrow}|\}$}
CreateClustersSkeleton(S) \tcp{Phase \rom{1} of the algorithm}
AssignWeakPoints(Q)  \tcp{Phase \rom{2} of the algorithm}

\caption{DenMune Algorithm}
\label{Alg:DenMune}
\end{algorithm}

\begin{algorithm}[H]
\setstretch{0.75}
\KwIn{ Sorted list $S$ of Seed points}
\KwOut{ Sorted list $L$ of the m generated clusters }
\tcp{Loop through all seed points and create clusters skeleton from seeds that share non-empty sets of MNN-seeds}

$i \leftarrow1$ \tcp {$i$ is a seed index}
$L \leftarrow \phi$ \tcp{List of clusters so far}
$ C  \leftarrow \cup(s_1, MNN({s_1}) )$ \\
$L$.append($C$)\\

$i \leftarrow i+1$  \tcp{increment i}

\ForEach{ $s_i \in S$ } {
    $C_{intersect} \leftarrow \phi$ \\
    $ C  \leftarrow \cup(s_i, MNN({s_i}) )$ \\
    \ForEach {$l \in L$} {
      \If {$l \cap C \neq \phi$ } {
        $C_{intersect} \leftarrow \cup (C_{intersect}, l)$ \\
       $L$.delete($l$)\\
         }
         }

       \If {$C_{intersect} \neq \phi$ } {  
    		$C \leftarrow \cup (C, C_{intersect})$ \\
    		$L$.append($C$)\\
    	}
     
     $i \leftarrow i+1$  \tcp{increment i}
  }
	$m \leftarrow Length(L)$ \\
	\For {$j$ from 1 to $m$}{
	$\ell(s \in C_j) \leftarrow j$ \tcp{label each seed point in $C_j$ as belonging to cluster $j$}
	}
\tcp{Output the set of generated clusters, $m$ the number of clusters and label each seed point $s$ belonging to a cluster $C_j$ by its corresponding cluster index} 
\caption{CreateClustersSkeleton(S)}
\label{Proc:CreateSkeleton}
\end{algorithm}

\begin{algorithm}[H]
\setstretch{0.75}
\KwIn{ Sorted lists $L$ of  m clusters and $Q$ of non-seed points.}
\KwOut{ Updated list $L$ of the m generated clusters.}
\tcp{Loop through all non-seed points and assign each of them to the cluster with which it shares the largest number of $MNN$-seeds}
$i \leftarrow 1$ \tcp  {$i$ is an index for non-seed points}
\ForEach{ $q_i \in Q$ }{
       Select $j$ such that  $|\{q_i \cup MNN_{q_i} \} \cap C_j |$ is maximum, where \hspace{1cm} $j=1,2,\cdots, m$ and $C_j \in L$\; 
       $\ell(q_i)  \leftarrow j$ \tcp{label non-seed point $q_i$ as belonging to cluster $C_j$}
       $i \leftarrow i+1$\;  
 }
 Output the formed clusters. The remaining unlabeled points are noise of type-2. 
 
\caption{AssignWeakPoints(Q)}
\label{Proc:AssignWeakPoints}
\end{algorithm}

\subsection{{Time Complexity}}
\label{subsec:time_complexity}
Given $N$ the number of data points, $K$ the number of nearest neighbors, $D$ the number of dimensions and $C$ the number of constructed clusters, the time complexity for computing the similarity matrix, between the data points, is $O(N^2)*D$ = $O(N^2)$, since $D =2$ (after dimensionality reduction). This complexity can be reduced to $O(N\log N)$, by the use of a data structure such as a k-d tree  \unskip~\cite{kd-tree-2013} and \unskip~\cite{optimized-quantization_2017}, which works efficiently with low dimensional data. The space complexity of this  preprocessing phase is $O(ND)$. The time complexity of the algorithm can be analyzed as 
follows:
\begin{itemize}
\item line 2, finding
$KNN_{p_i\rightarrow}$: needs K iterations for each data point, hence it has a complexity of $O(NK)$ 
\item lines 3-6, finding $KNN_{p_i\leftarrow}$: needs K iterations for each of the N data points, hence it has a complexity of $O(KN)$
\item line 7, finding $MNN$ for each of the N data points, a search for mutual neighborhood is done within the K-nearest neighbors of each point. \item line 9, sorting points: has a complexity of $O(N\log N)$, using binary sort.
\item CreateClustersSkeleton algorithm
has a complexity of $O(|S|*|R|* \log K)$, where R is an upper bound on the number of temporarily generated clusters, $m \le R \le |S|$. Letting O(R) $\approx |S|$ and O(|S|) $\approx N$, then this complexity becomes $\approx O(N^2 \log K)$.
 \item Similarly AssignWeakPoints algorithm has a complexity of $O(|Q| *|R|* K)$, since we iterate through each of the $Q$ weak data points, searching for the maximum intersection between its MNN and each of the formed clusters. Therefore, this complexity becomes $\approx O(N^2 K)$.
\end {itemize}

The overall time complexity for DenMune algorithm is O($N^2K$) and its space complexity is $O(NK)$.

\section{Experimental Results}
\label{section:EXPERIMENTAL_RESULTS}

\begingroup
\setlength{\tabcolsep}{8pt} 
\renewcommand{\arraystretch}{0.7} 
\centering

\begin{table}[H]
\caption{Datasets used in the experiments and their properties}
\label{Tab:datasets}
\centering
\resizebox{\textwidth}{!}{%
\begin{tabular}{lcccc}
\hline
Dataset          & Type      & Size & Number of dimensions & Number of clusters \\
\hline
\hline
A1               & Synthetic & 3000    & 2          & 20       \\
A2               & Synthetic & 2050    & 2          & 35       \\
Aggregation      & Synthetic & 788     & 2          & 7        \\
Compound         & Synthetic & 399     & 2          & 6        \\
D31              & Synthetic & 3100    & 2          & 31       \\
Dim-32           & Synthetic & 1024    & 32         & 16       \\
Dim-128          & Synthetic & 1024    & 128        & 16       \\
Dim-512          & Synthetic & 1024    & 512        & 16       \\
Flame            & Synthetic & 240     & 2          & 2        \\
G2-2-10          & Synthetic & 2048    & 2          & 2        \\
G2-2-30          & Synthetic & 2048    & 2          & 2        \\
G2-2-50          & Synthetic & 2048    & 2          & 2        \\
Jain             & Synthetic & 373     & 2          & 2        \\
Mouse            & Synthetic & 500     & 2          & 3        \\
Pathbased        & Synthetic & 300     & 2          & 3        \\
R15              & Synthetic & 600     & 2          & 15       \\
S1               & Synthetic & 5000    & 2          & 15       \\
S2               & Synthetic & 5000    & 2          & 15       \\
Spiral           & Synthetic & 312     & 2          & 3        \\
Unbalance        & Synthetic & 6500    & 2          & 8        \\
Vary density     & Synthetic & 150     & 2          & 3        \\
Appendicitis     & Real      & 106     & 7          & 2        \\
Arcene           & Real      & 200     & 10000      & 2        \\
Breast cancer    & Real      & 683     & 9          & 2        \\
Optical digits   & Real      & 5620    & 64         & 10       \\
Pendigits        & Real      & 10992   & 16         & 10       \\
Ecoli            & Real      & 336     & 8          & 8        \\
Glass            & Real      & 214     & 9          & 6        \\
Iris             & Real      & 150     & 4          & 3        \\
MNIST            & Real      & 70000   & 784        & 10       \\
Libras movement  & Real      & 360     & 91         & 15       \\
Robot navigation & Real      & 5456    & 24         & 4        \\
SCC              & Real      & 600     & 60         & 6        \\
Seeds            & Real      & 210     & 7          & 3        \\
WDBC             & Real      & 569     & 32         & 2        \\
Yeast            & Real      & 1484    & 8          & 10      \\
\hline
\end{tabular}%
}
\end{table}
\endgroup

We have conducted extensive experiments on the datasets described in Table \ref{Tab:datasets} which include: 
(1) Fifteen real datasets obtained from UCI repository \ucidatasets{}, MNIST dataset\mnistdatasets{} and KEEL datasets\keeldatasets{} (2) Twenty-one synthetic datasets from \benchmarkdatasets{} and \elkidatasets{}. In total, thirty-six datasets have been used to assess the results obtained by DenMune with respect to the ground truth as well as to the results obtained by nine known algorithms, NPIR \unskip~\cite{npir_2020},  CBKM  \unskip~\cite{cbkm_2019}, Fast DP \unskip~\cite{fastdp_2019}, FINCH  \unskip~\cite{finch_2019}), RS \unskip~\cite{rs_2018}), RCC \unskip~\cite{rcc_2017}) HDBSCAN \unskip~\cite{hdbscan_2017}, KMeans++ \unskip~\cite{kmeanspp_2007} and Spectral clustering.

The Euclidean distance has been adopted as a similarity metric for all datasets.

\subsection {Dimensionality Reduction}
Datasets often contain a large number of features, which may even outnumber the observations as in the Arcene dataset. Due to the computational and theoretical challenges associated with high dimensional data, reducing the dimension while maintaining the structure of the original data is desirable \unskip~\cite{dimensionality_reduction}. 
Also, high dimensional data may contain many irrelevant dimensions that suppress each others. These issues can confuse any clustering algorithm by hiding clusters, especially in noisy data. For these reasons, all datasets have been reduced to two dimensions, using the t-sne algorithm  \unskip~\cite{tsne-2014}, before applying the examined algorithms on them. DenMune has been examined on the ten real datasets listed in Table \ref{Tab:datasets}, using various dimensionality  reduction techniques. In general, as shown in Table \ref{tab:2d_nd}, the algorithm performance on  a dataset projected to 2-D is better than its performance on the same dataset in its high dimension version. Table \ref{tab:reduction_alg} shows that  DenMune attains its best performance when t-sne is used for dimensionality reduction. t-sne outperforms Principal Component Analysis (PCA), Factor Analysis (FA) and Non-negative Matrix Factorization (NMF) by a large margin.


\subsection{Agorithms' Implementation and Parameters' setting}
For HDBSCAN, Spectral Clustering and Kmeans++  the implementations provided by SKlearn\sklearn have been adopted. All other algorithms, NBIR, CBKM, RS, FINCH, FastDP and RCC implementations are provided by authors themselves. DenMune algorithm has been implemented in C++ and integrated with SKlearn, to benefit from its libraries in computing various validation indexes.
 
The parameters for each algorithm have been selected according to each algorithm defaults and recommendations. (1) for NPIR, the IR parameter is selected in the range [0.01, 0.05, 0.10, 0.15, 0.20], with  ten iterations for each run. (2) for HDBSCAN, The primary parameter and the most intuitive parameter is  min-cluster-size is selected in the range [2..100], (3) for Spectral clustering  and KMeans++ the default parameters in SKlearn have been adopted. The number of clusters is set equal to the ground truth. Each algorithm is run 100 times for each dataset and the best performance is recorded, (4) for DenMune, the only used parameter, $K$ is selected in the range [1..50] for small datasets and [1..200] for big datasets. For MNIST dataset, NPIR failed to scale to adapt to this big dataset even on a cloud server with 128 GB memory, thus all MNIST results were removed from the ranking process for all other algorithms.

\subsection{Results and Discussion}
The twenty-one synthetic datasets, listed in Table \ref{Tab:datasets}, have been used to demonstrate the efficiency of our proposed algorithm. All datasets are 2-D  except DIM datasets which are reduced from 32, 128 and 512 to 2-D to make them easy to visualize.  They are of different sizes  (G2 and DIM datasets). They have clusters of different densities, shapes (Spiral, Compound,  Flame and Pathbased datasets) and degree of overlapping (S1 and G2 datasets). Revealing the inherent structure of these datasets is challenging for most heuristic algorithms. Three metrics have been  recorded (1) The  F1 scores are recorded in Tables \ref{Tab:f1_synthetic} and \ref{Tab:f1_real} for synthetic and real datasets, respectively. The Normalized Mutual Information, NMI is recorded in Tables \ref{Tab:nmi-synthetic} and \ref{Tab:nmi-real} for synthetic and real datasets, respectively, and the Adjusted Rand Index, ARI is recorded in Tables \ref{Tab:ari-synthetic} and \ref{Tab:ari-real} for synthetic and real datasets, respectively.

We adopt a ranking system to order algorithms , based on their clustering performance, as measured by F1, NMI and ARI scores. The lower the rank of an algorithm, the better its clustering quality for the datasets examined.
Three ranking values are added to the bottom of Tables (\ref{Tab:f1_synthetic} : \ref{Tab:ari-real}) as follows: (1) Total rank: sum of the ranks of an algorithm over the examined datasets (2) Average rank: Total rank divided by the number of datasets and (3) rank: the algorithm ranking among the set of examined algorithms, given in ascending order (lower ranks preferred).

\begin{figure}[H]
   
  \begin{minipage}[b]{\linewidth}
   \centering
   
      \begin{subfigure}[t]{.24\textwidth}
    \centering
     \includegraphics[width=\linewidth, height=2.3 cm]{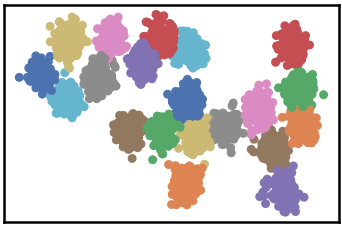}
      \vspace*{-1.0cm}
     \caption{A1 dataset}\label{Fig:ground_a1}
     \end{subfigure}\hfill
   \begin{subfigure}[t]{.24\textwidth}
     \centering
     \includegraphics[width=\linewidth, height=2.3 cm]{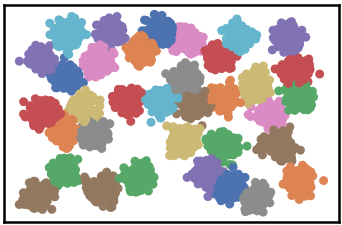}
      \vspace*{-1.0cm}
     \caption{A2 dataset}\label{Fig:ground_a2}
   \end{subfigure}\hfill
   \begin{subfigure}[t]{.24\textwidth}
     \centering
     \includegraphics[width=\linewidth, height=2.3 cm]{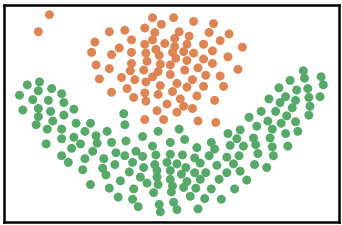}
      \vspace*{-1.0cm}
     \caption{Flame dataset}\label{Fig:ground_flame}
   \end{subfigure}\hfill
   \begin{subfigure}[t]{.24\textwidth}
     \centering
     \includegraphics[width=\linewidth, height=2.3 cm]{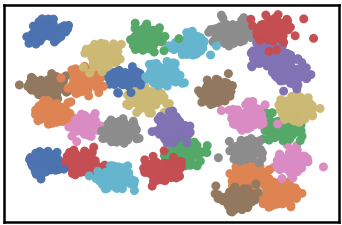}
      \vspace*{-1.0cm}
     \caption{D31 dataset}\label{Fig:ground_d31}
   \end{subfigure}\hfill
  
   \begin{subfigure}[t]{.24\textwidth}
    \centering
     \includegraphics[width=\linewidth, height=2.3 cm]{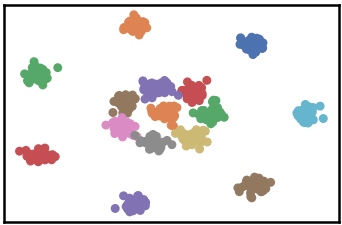}
      \vspace*{-1.0cm}
     \caption{R15 dataset}\label{Fig:ground_r15}
     \end{subfigure}\hfill
   \begin{subfigure}[t]{.24\textwidth}
     \centering
     \includegraphics[width=\linewidth, height=2.3 cm]{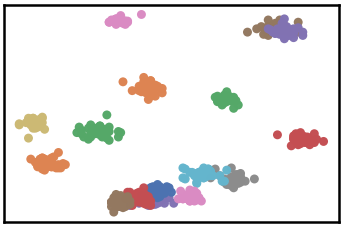}
      \vspace*{-1.0cm}
     \caption{Dim-32 dataset}\label{Fig:ground_dim032}
   \end{subfigure}\hfill
   \begin{subfigure}[t]{.24\textwidth}
     \centering
     \includegraphics[width=\linewidth, height=2.3 cm]{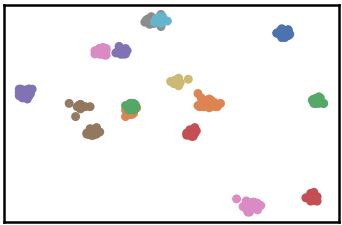}
      \vspace*{-1.0cm}
     \caption{Dim-128 dataset}\label{Fig:ground_dim128}
   \end{subfigure}\hfill
   \begin{subfigure}[t]{.24\textwidth}
     \centering
     \includegraphics[width=\linewidth, height=2.3 cm]{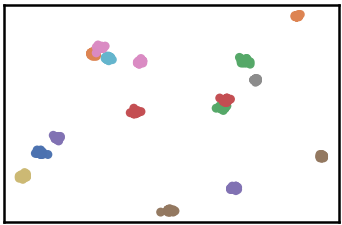}
      \vspace*{-1.0cm}
     \caption{Dim-512 dataset}\label{Fig:ground_dim512}
   \end{subfigure}\hfill

  \begin{subfigure}[t]{.24\textwidth}
    \centering
     \includegraphics[width=\linewidth, height=2.3 cm]{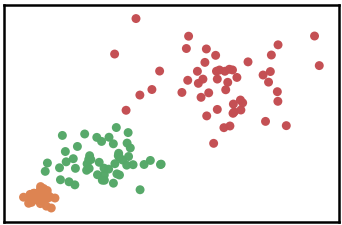}
     \vspace*{-1.0cm}
     \caption{Vary-density dataset}\label{Fig:ground_varydensity}
     \end{subfigure}\hfill
   \begin{subfigure}[t]{.24\textwidth}
     \centering
     \includegraphics[width=\linewidth, height=2.3 cm]{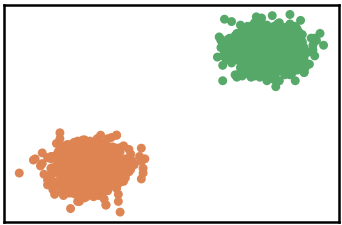}
      \vspace*{-1.0cm}
     \caption{G2-2-10 dataset}\label{Fig:ground_g2-2-10}
   \end{subfigure}\hfill
   \begin{subfigure}[t]{.24\textwidth}
     \centering
     \includegraphics[width=\linewidth, height=2.3 cm]{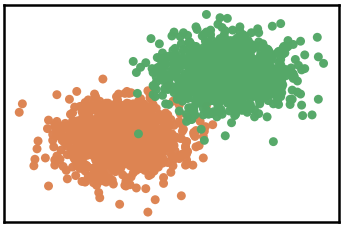}
      \vspace*{-1.0cm}
     \caption{G2-2-30 dataset}\label{Fig:ground_g2-2-30}
   \end{subfigure}\hfill
   \begin{subfigure}[t]{.24\textwidth}
     \centering
     \includegraphics[width=\linewidth, height=2.3 cm]{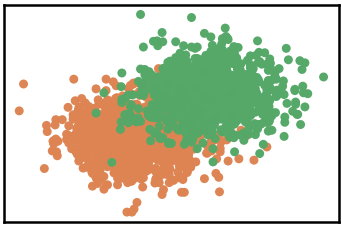}
      \vspace*{-1.0cm}
     \caption{G2-2-50 dataset}\label{Fig:ground_g2-2-50}
   \end{subfigure}\hfill

  \begin{subfigure}[t]{.24\textwidth}
    \centering
     \includegraphics[width=\linewidth, height=2.3 cm]{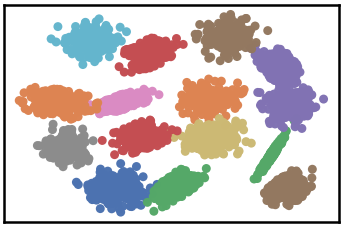}
     \vspace*{-1.0cm}
     \caption{S1 dataset}\label{Fig:ground_s1}
     \end{subfigure}\hfill
   \begin{subfigure}[t]{.24\textwidth}
     \centering
     \includegraphics[width=\linewidth, height=2.3 cm]{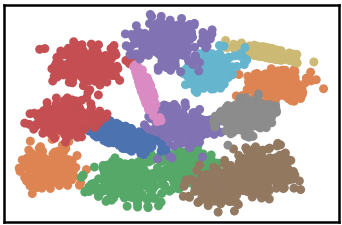}
      \vspace*{-1.0cm}
     \caption{S2 dataset}\label{Fig:ground_s2}
   \end{subfigure}\hfill
   \begin{subfigure}[t]{.24\textwidth}
     \centering
     \includegraphics[width=\linewidth, height=2.3 cm]{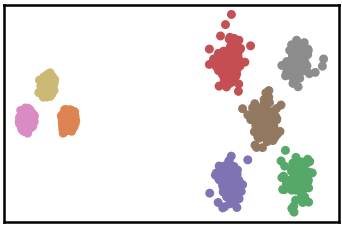}
      \vspace*{-1.0cm}
     \caption{Unbalance dataset}\label{Fig:ground_unbalance}
   \end{subfigure}\hfill
   \begin{subfigure}[t]{.24\textwidth}
     \centering
     \includegraphics[width=\linewidth, height=2.3 cm]{images/visualize/empty.png}
      \vspace*{-1.0cm}
    
   \end{subfigure}\hfill

   \end{minipage}\hfill
    \vspace*{-0.3cm}
   \caption{Ground truths of the fifteen synthetic datasets, visualized using t-sne, and used for algorithms' comparisons.}
   \label{Fig:ground_synthetic}

   \end{figure}

\begin{figure}[H]
   
  \begin{minipage}[b]{\linewidth}
   \centering
   
      \begin{subfigure}[t]{.24\textwidth}
    \centering
     \includegraphics[width=\linewidth, height=2.3 cm]{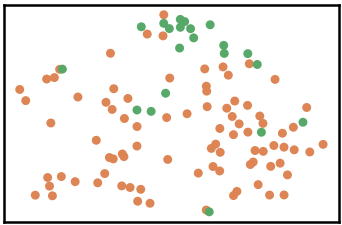}
      \vspace*{-1.0cm}
     \caption{Appendicitis
 dataset}\label{Fig:appcity}
     \end{subfigure}\hfill
   \begin{subfigure}[t]{.24\textwidth}
     \centering
     \includegraphics[width=\linewidth, height=2.3 cm]{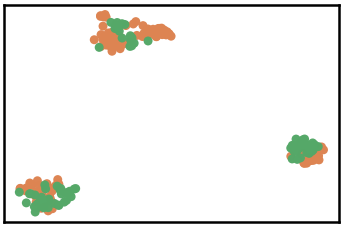}
      \vspace*{-1.0cm}
     \caption{Arcene
 dataset}\label{Fig:ground_arcene}
   \end{subfigure}\hfill
   \begin{subfigure}[t]{.24\textwidth}
     \centering
     \includegraphics[width=\linewidth, height=2.3 cm]{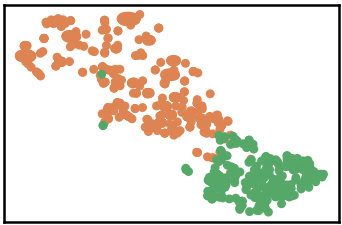}
      \vspace*{-1.0cm}
     \caption{Breast-cancer
 dataset}\label{Fig:ground_breast}
   \end{subfigure}\hfill
   \begin{subfigure}[t]{.24\textwidth}
     \centering
     \includegraphics[width=\linewidth, height=2.3 cm]{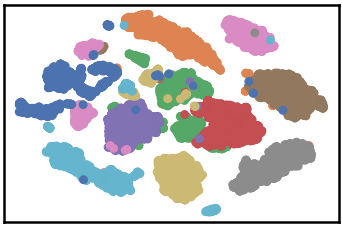}
      \vspace*{-1.0cm}
     \caption{Pendigits
 dataset}\label{Fig:ground_digits}
   \end{subfigure}\hfill
  
   \begin{subfigure}[t]{.24\textwidth}
    \centering
     \includegraphics[width=\linewidth, height=2.3 cm]{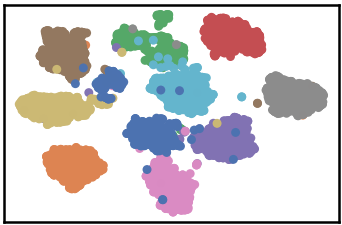}
      \vspace*{-1.0cm}
     \caption{Optical digits
 dataset}\label{Fig:ground_chars}
     \end{subfigure}\hfill
   \begin{subfigure}[t]{.24\textwidth}
     \centering
     \includegraphics[width=\linewidth, height=2.3 cm]{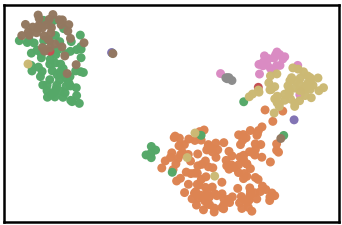}
      \vspace*{-1.0cm}
     \caption{Ecoli dataset}\label{Fig:ground_ecoli}
   \end{subfigure}\hfill
   \begin{subfigure}[t]{.24\textwidth}
     \centering
     \includegraphics[width=\linewidth, height=2.3 cm]{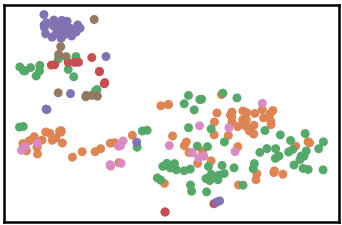}
      \vspace*{-1.0cm}
     \caption{Glass dataset}\label{Fig:ground_glass}
   \end{subfigure}\hfill
   \begin{subfigure}[t]{.24\textwidth}
     \centering
     \includegraphics[width=\linewidth, height=2.3 cm]{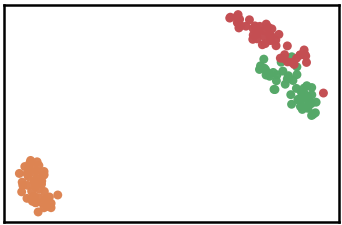}
      \vspace*{-1.0cm}
     \caption{iris dataset}\label{Fig:ground_iris}
   \end{subfigure}\hfill

  \begin{subfigure}[t]{.24\textwidth}
    \centering
     \includegraphics[width=\linewidth, height=2.3 cm]{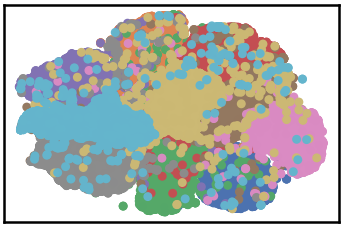}
     \vspace*{-1.0cm}
     \caption{MNIST dataset}\label{Fig:ground_mnist}
     \end{subfigure}\hfill
   \begin{subfigure}[t]{.24\textwidth}
     \centering
     \includegraphics[width=\linewidth, height=2.3 cm]{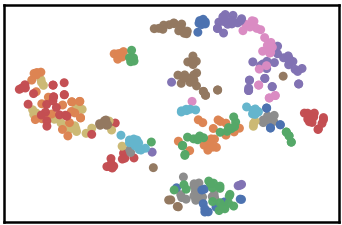}
      \vspace*{-1.0cm}
     \caption{Libras move.
 dataset}\label{Fig:ground_move}
   \end{subfigure}\hfill
   \begin{subfigure}[t]{.24\textwidth}
     \centering
     \includegraphics[width=\linewidth, height=2.3 cm]{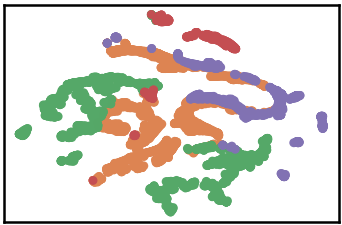}
      \vspace*{-1.0cm}
     \caption{Robot nav. dataset}\label{Fig:ground_robot2d}
   \end{subfigure}\hfill
   \begin{subfigure}[t]{.24\textwidth}
     \centering
     \includegraphics[width=\linewidth, height=2.3 cm]{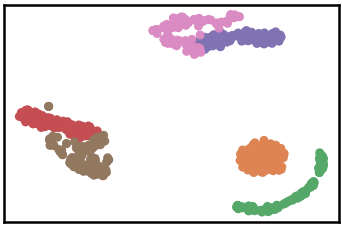}
      \vspace*{-1.0cm}
     \caption{SCC dataset}\label{Fig:ground_scc}
   \end{subfigure}\hfill

  \begin{subfigure}[t]{.24\textwidth}
    \centering
     \includegraphics[width=\linewidth, height=2.3 cm]{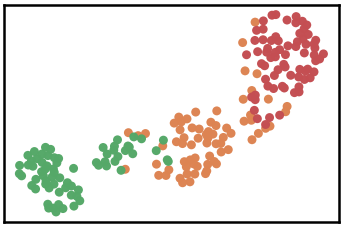}
     \vspace*{-1.0cm}
     \caption{Seeds dataset}\label{Fig:ground_seeds}
     \end{subfigure}\hfill
   \begin{subfigure}[t]{.24\textwidth}
     \centering
     \includegraphics[width=\linewidth, height=2.3 cm]{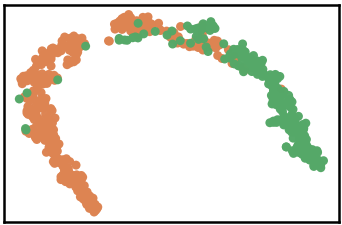}
      \vspace*{-1.0cm}
     \caption{WDBC dataset}\label{Fig:ground_wdbc}
   \end{subfigure}\hfill
   \begin{subfigure}[t]{.24\textwidth}
     \centering
     \includegraphics[width=\linewidth, height=2.3 cm]{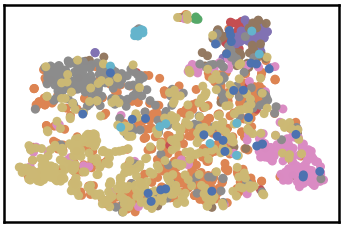}
      \vspace*{-1.0cm}
     \caption{Yeast dataset}\label{Fig:ground_yeast}
   \end{subfigure}\hfill
   \begin{subfigure}[t]{.24\textwidth}
     \centering
     \includegraphics[width=\linewidth, height=2.3 cm]{images/visualize/empty.png}
      \vspace*{-1.0cm}
    
   \end{subfigure}\hfill

   \end{minipage}\hfill
    \vspace*{-0.3cm}
   \caption{Ground truths of the fifteen real datasets, visualized using t-sne, and used for algorithm's comparisons.}
   \label{Fig:ground_real}

   \end{figure}

The ground truth for all datasets are visualised using the t-sne algorithm as shown in Fig. (\ref{Fig:ground_synthetic} and \ref{Fig:ground_real}) for synthetic and real datasets, respectively.

In general, the results show that DenMune outperforms all other algorithms for the majority of the datasets examined. Denmune has the lowest rank values for each of the three validity indexes used in the assessment for both synthetic, Tables(  \ref{Tab:f1_synthetic},  \ref{Tab:nmi-synthetic} and  \ref{Tab:ari-synthetic}) and real datasets, Tables(  \ref{Tab:f1_real},  \ref{Tab:nmi-real} and  \ref{Tab:ari-real}).

Based on F1-score, Denmune outperforms the other algorithms for twenty-eight out of the thirty-six datasets. For the remaining datasets.

(1) Arcene dataset: all algorithms (except for Finch and RCC algorithms) outperform Denmune (+8\%). 
(2) G2-2-50 dataset: CBKM, RS and FastDP algorithms outperform DenMune (+2\%).
(3) Iris dataset: NPIR, CBKM, RS and Spectral outperform DenMune (+1\% : +8\%).
(4) Glass dataset: Spectral Clustering outperforms DenMune slightly (+2\%).
(5) SCC dataset: RS algorithm outperforms all other algorithms for this dataset with noticable F1-score (84\%) then comes RCC with (77\%), while Denmune scores only 68\%.
(6) Seeds dataset: NPIR and CBKM outperform DenMune (+1\% : +2\%).
(7). WDBC dataset: NPIR, RS and FastDP outperform DenMune (+2\% : +7\%)
(8) Yeast dataset: CBKM, FINCH, RCC, KMeans++ outperform DenMune (+1\% : +5\%).

We are going to investigate why DenMune outperform for the majority of datasets, then we will ilusterate why some algorithms outperform DenMune for some datasets.

DenMune has a noticeable better performance over other algorithms for many datasets such as (1) A1 dataset (+33\%), (2) A2 dataset (+14\%), (3) Compound dataset (+8\%), (4) D31 dataset  (+34\%), (5) Dim-128 dataset (+20\%), (6) Pathbased dataset (+10\%), (7) S1 dataset (+22\%) ,(8) S2 datset (+25\%) , (9) Optical-digits dataset (+27\%) , (10) Pendigits dataset (+24\%), (11) Ecoli dataset (+6\%) and  (12) MNIST datasets (+6\%). This is due to the framework DenMune adopts in its clustering process which allows it to distinguish real clusters in noisy data, even if they are attached to each others or overlapped as long as they are of distinguishable densities.

\begin{figure}[H]
   
  \begin{minipage}[b]{\linewidth}
   \centering
   
      \begin{subfigure}[t]{.24\textwidth}
    \centering
     \includegraphics[width=\linewidth, height=2.3 cm]{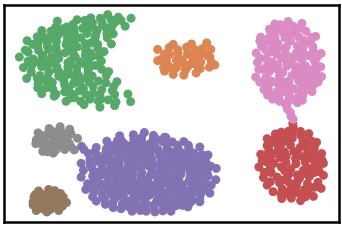}
      \vspace*{-1.0cm}
     \caption{Aggregation dataset}\label{Fig:ground_aggregation}
     \end{subfigure}\hfill
   \begin{subfigure}[t]{.24\textwidth}
     \centering
     \includegraphics[width=\linewidth, height=2.3 cm]{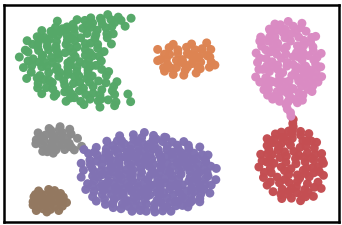}
      \vspace*{-1.0cm}
     \caption{DenMune}\label{Fig:denmune_aggregation}
   \end{subfigure}\hfill
   \begin{subfigure}[t]{.24\textwidth}
     \centering
     \includegraphics[width=\linewidth, height=2.3 cm]{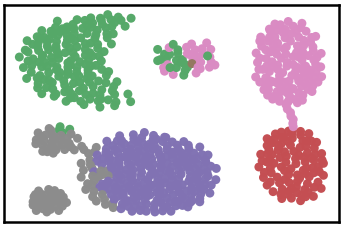}
      \vspace*{-1.0cm}
     \caption{NPIR}\label{Fig:npir_aggregation}
   \end{subfigure}\hfill
   \begin{subfigure}[t]{.24\textwidth}
     \centering
     \includegraphics[width=\linewidth, height=2.3 cm]{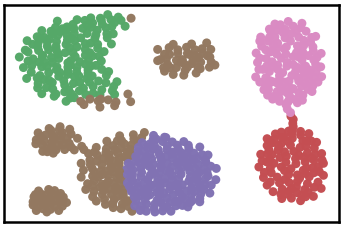}
      \vspace*{-1.0cm}
     \caption{CBKM}\label{Fig:cbkm_aggregation}
   \end{subfigure}\hfill
  
   \begin{subfigure}[t]{.24\textwidth}
    \centering
     \includegraphics[width=\linewidth, height=2.3 cm]{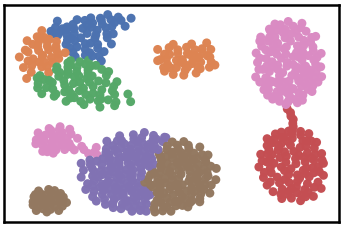}
      \vspace*{-1.0cm}
     \caption{FINCH}\label{Fig:finch_aggregation}
     \end{subfigure}\hfill
   \begin{subfigure}[t]{.24\textwidth}
     \centering
     \includegraphics[width=\linewidth, height=2.3 cm]{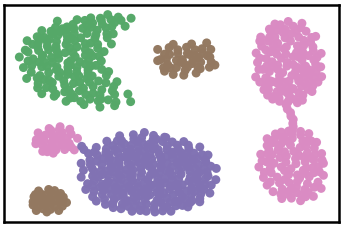}
      \vspace*{-1.0cm}
     \caption{FastDP}\label{Fig:fastdp_aggregation}
   \end{subfigure}\hfill
   \begin{subfigure}[t]{.24\textwidth}
     \centering
     \includegraphics[width=\linewidth, height=2.3 cm]{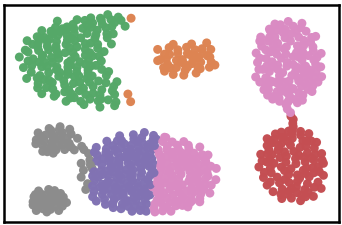}
      \vspace*{-1.0cm}
     \caption{RS}\label{Fig:rs_aggregation}
   \end{subfigure}\hfill
   \begin{subfigure}[t]{.24\textwidth}
     \centering
     \includegraphics[width=\linewidth, height=2.3 cm]{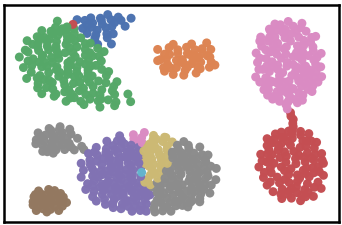}
      \vspace*{-1.0cm}
     \caption{RCC}\label{Fig:rcc_aggregation}
   \end{subfigure}\hfill

  \begin{subfigure}[t]{.24\textwidth}
    \centering
     \includegraphics[width=\linewidth, height=2.3 cm]{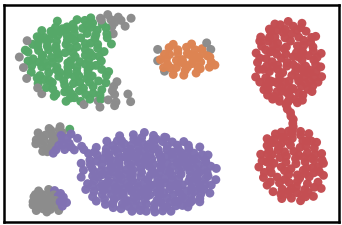}
     \vspace*{-1.0cm}
     \caption{HDBSCAN}\label{Fig:hdbscan_aggregation}
     \end{subfigure}\hfill
   \begin{subfigure}[t]{.24\textwidth}
     \centering
     \includegraphics[width=\linewidth, height=2.3 cm]{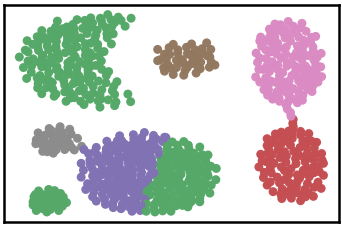}
      \vspace*{-1.0cm}
     \caption{Spectral clustering}\label{Fig:spectral_aggregation}
   \end{subfigure}\hfill
   \begin{subfigure}[t]{.24\textwidth}
     \centering
     \includegraphics[width=\linewidth, height=2.3 cm]{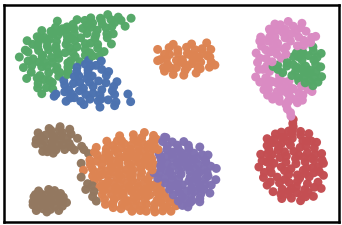}
      \vspace*{-1.0cm}
     \caption{KMeans++}\label{Fig:kmeanspp_aggregation}
   \end{subfigure}\hfill
   \begin{subfigure}[t]{.24\textwidth}
     \centering
     \includegraphics[width=\linewidth, height=2.3 cm]{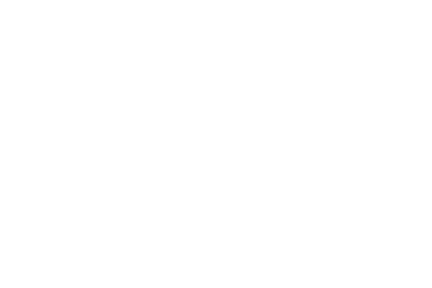}
      \vspace*{-1.0cm}
    
   \end{subfigure}\hfill

   \end{minipage}\hfill
    \vspace*{-0.3cm}
   \caption{Visualization of the results obtained by the ten algorithms for the Aggregation dataset.}
   \label{Fig:Aggregation_Results}

   \end{figure}

\begin{figure}[H]
   
  \begin{minipage}[b]{\linewidth}
   \centering
   
      \begin{subfigure}[t]{.24\textwidth}
    \centering
     \includegraphics[width=\linewidth, height=2.3 cm]{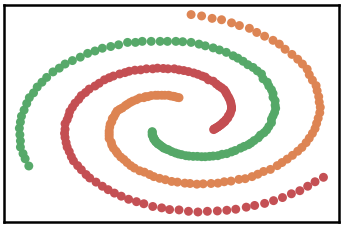}
      \vspace*{-1.0cm}
     \caption{Spiral dataset}\label{Fig:ground_spiral}
     \end{subfigure}\hfill
   \begin{subfigure}[t]{.24\textwidth}
     \centering
     \includegraphics[width=\linewidth, height=2.3 cm]{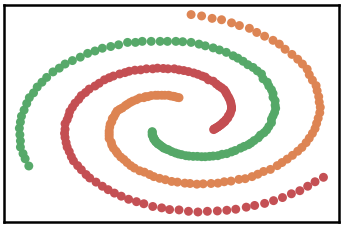}
      \vspace*{-1.0cm}
     \caption{DenMune}\label{Fig:denmune_spiral}
   \end{subfigure}\hfill
   \begin{subfigure}[t]{.24\textwidth}
     \centering
     \includegraphics[width=\linewidth, height=2.3 cm]{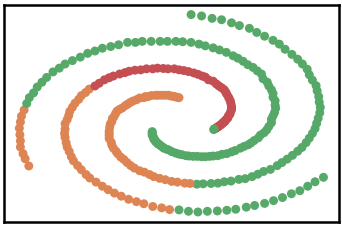}
      \vspace*{-1.0cm}
     \caption{NPIR}\label{Fig:npir_spiral}
   \end{subfigure}\hfill
   \begin{subfigure}[t]{.24\textwidth}
     \centering
     \includegraphics[width=\linewidth, height=2.3 cm]{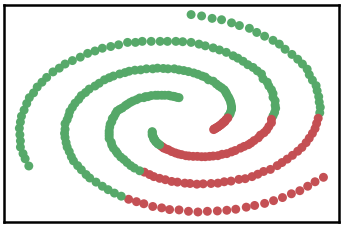}
      \vspace*{-1.0cm}
     \caption{CBKM}\label{Fig:cbkm_spiral}
   \end{subfigure}\hfill
  
   \begin{subfigure}[t]{.24\textwidth}
    \centering
     \includegraphics[width=\linewidth, height=2.3 cm]{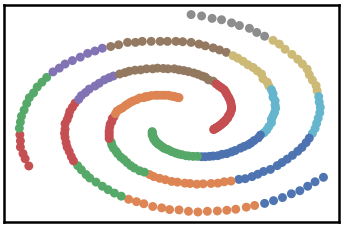}
      \vspace*{-1.0cm}
     \caption{FINCH}\label{Fig:finch_spiral}
     \end{subfigure}\hfill
   \begin{subfigure}[t]{.24\textwidth}
     \centering
     \includegraphics[width=\linewidth, height=2.3 cm]{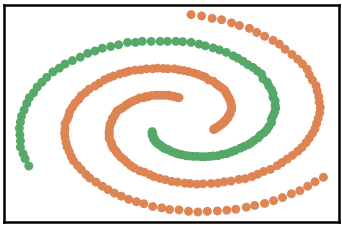}
      \vspace*{-1.0cm}
     \caption{FastDP}\label{Fig:fastdp_spiral}
   \end{subfigure}\hfill
   \begin{subfigure}[t]{.24\textwidth}
     \centering
     \includegraphics[width=\linewidth, height=2.3 cm]{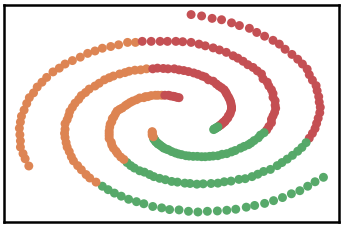}
      \vspace*{-1.0cm}
     \caption{RS}\label{Fig:rs_spiral}
   \end{subfigure}\hfill
   \begin{subfigure}[t]{.24\textwidth}
     \centering
     \includegraphics[width=\linewidth, height=2.3 cm]{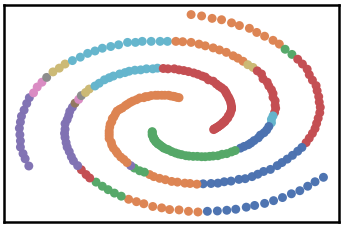}
      \vspace*{-1.0cm}
     \caption{RCC}\label{Fig:rcc_spiral}
   \end{subfigure}\hfill

  \begin{subfigure}[t]{.24\textwidth}
    \centering
     \includegraphics[width=\linewidth, height=2.3 cm]{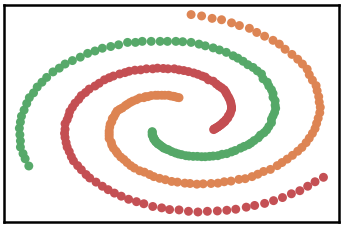}
     \vspace*{-1.0cm}
     \caption{HDBSCAN}\label{Fig:hdbscan_spiral}
     \end{subfigure}\hfill
   \begin{subfigure}[t]{.24\textwidth}
     \centering
     \includegraphics[width=\linewidth, height=2.3 cm]{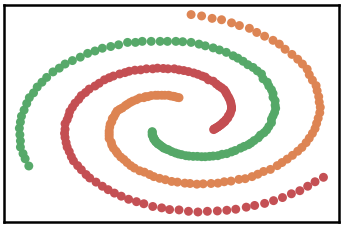}
      \vspace*{-1.0cm}
     \caption{Spectral clustering}\label{Fig:spectral_spiral}
   \end{subfigure}\hfill
   \begin{subfigure}[t]{.24\textwidth}
     \centering
     \includegraphics[width=\linewidth, height=2.3 cm]{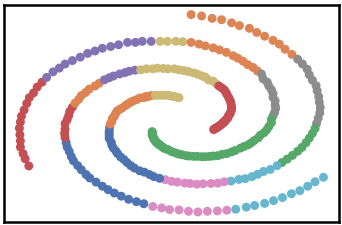}
      \vspace*{-1.0cm}
     \caption{KMeans++}\label{Fig:kmeanspp_spiral}
   \end{subfigure}\hfill
   \begin{subfigure}[t]{.24\textwidth}
     \centering
     \includegraphics[width=\linewidth, height=2.3 cm]{images/visualize/empty.png}
      \vspace*{-1.0cm}
    
   \end{subfigure}\hfill

   \end{minipage}\hfill
    \vspace*{-0.3cm}
   \caption{Visualization of the results obtained by the ten algorithms for the Spiral dataset.}
   \label{Fig:Spiral_Results}

   \end{figure}

\begin{figure}[H]
   
  \begin{minipage}[b]{\linewidth}
   \centering
   
      \begin{subfigure}[t]{.24\textwidth}
    \centering
     \includegraphics[width=\linewidth, height=2.3 cm]{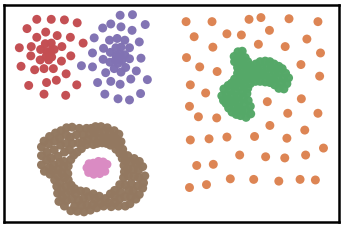}
      \vspace*{-1.0cm}
     \caption{Compound dataset}\label{Fig:ground_compound}
     \end{subfigure}\hfill
   \begin{subfigure}[t]{.24\textwidth}
     \centering
     \includegraphics[width=\linewidth, height=2.3 cm]{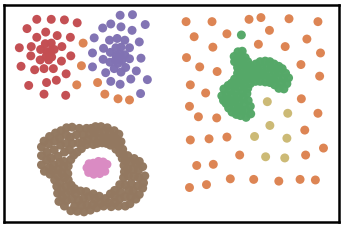}
      \vspace*{-1.0cm}
     \caption{DenMune}\label{Fig:denmune_compound}
   \end{subfigure}\hfill
   \begin{subfigure}[t]{.24\textwidth}
     \centering
     \includegraphics[width=\linewidth, height=2.3 cm]{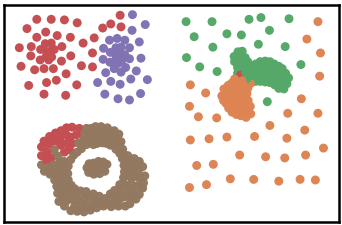}
      \vspace*{-1.0cm}
     \caption{NPIR}\label{Fig:npir_compound}
   \end{subfigure}\hfill
   \begin{subfigure}[t]{.24\textwidth}
     \centering
     \includegraphics[width=\linewidth, height=2.3 cm]{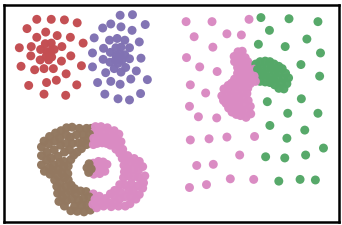}
      \vspace*{-1.0cm}
     \caption{CBKM}\label{Fig:cbkm_compound}
   \end{subfigure}\hfill
  
   \begin{subfigure}[t]{.24\textwidth}
    \centering
     \includegraphics[width=\linewidth, height=2.3 cm]{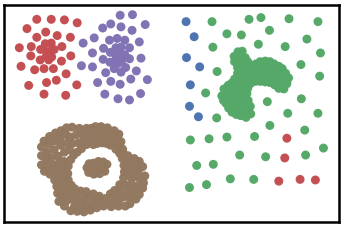}
      \vspace*{-1.0cm}
     \caption{FINCH}\label{Fig:finch_compound}
     \end{subfigure}\hfill
   \begin{subfigure}[t]{.24\textwidth}
     \centering
     \includegraphics[width=\linewidth, height=2.3 cm]{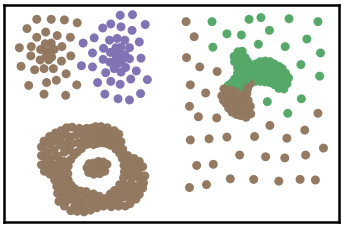}
      \vspace*{-1.0cm}
     \caption{FastDP}\label{Fig:fastdp_compound}
   \end{subfigure}\hfill
   \begin{subfigure}[t]{.24\textwidth}
     \centering
     \includegraphics[width=\linewidth, height=2.3 cm]{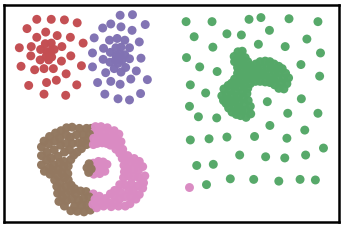}
      \vspace*{-1.0cm}
     \caption{RS}\label{Fig:rs_compound}
   \end{subfigure}\hfill
   \begin{subfigure}[t]{.24\textwidth}
     \centering
     \includegraphics[width=\linewidth, height=2.3 cm]{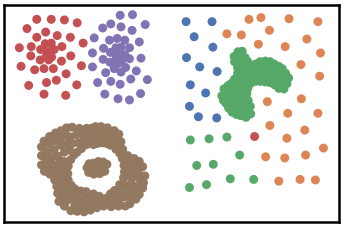}
      \vspace*{-1.0cm}
     \caption{RCC}\label{Fig:rcc_compound}
   \end{subfigure}\hfill

  \begin{subfigure}[t]{.24\textwidth}
    \centering
     \includegraphics[width=\linewidth, height=2.3 cm]{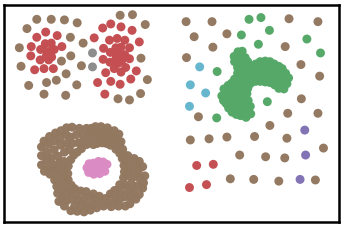}
     \vspace*{-1.0cm}
     \caption{HDBSCAN}\label{Fig:hdbscan_compound}
     \end{subfigure}\hfill
   \begin{subfigure}[t]{.24\textwidth}
     \centering
     \includegraphics[width=\linewidth, height=2.3 cm]{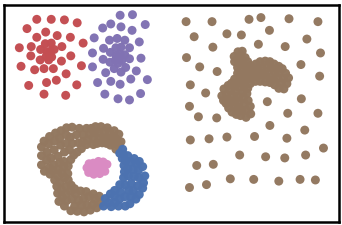}
      \vspace*{-1.0cm}
     \caption{Spectral clustering}\label{Fig:spectral_compound}
   \end{subfigure}\hfill
   \begin{subfigure}[t]{.24\textwidth}
     \centering
     \includegraphics[width=\linewidth, height=2.3 cm]{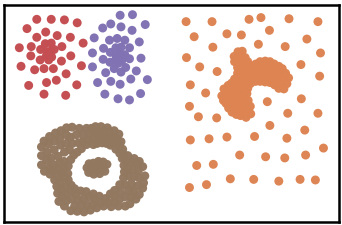}
      \vspace*{-1.0cm}
     \caption{KMeans++}\label{Fig:kmeanspp_compound}
   \end{subfigure}\hfill
   \begin{subfigure}[t]{.24\textwidth}
     \centering
     \includegraphics[width=\linewidth, height=2.3 cm]{images/visualize/empty.png}
      \vspace*{-1.0cm}
    
   \end{subfigure}\hfill

   \end{minipage}\hfill
    \vspace*{-0.3cm}
   \caption{Visualization of the results obtained by the ten algorithms for the  Compound dataset.}
   \label{Fig:Compound_Results}

   \end{figure}

\begin{figure}[H]
   
  \begin{minipage}[b]{\linewidth}
   \centering
   
      \begin{subfigure}[t]{.24\textwidth}
    \centering
     \includegraphics[width=\linewidth, height=2.3 cm]{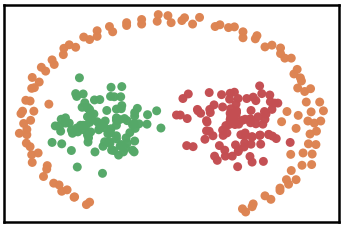}
      \vspace*{-1.0cm}
     \caption{Pathbased dataset}\label{Fig:ground_pathbased}
     \end{subfigure}\hfill
   \begin{subfigure}[t]{.24\textwidth}
     \centering
     \includegraphics[width=\linewidth, height=2.3 cm]{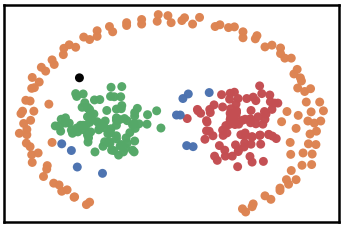}
      \vspace*{-1.0cm}
     \caption{DenMune}\label{Fig:denmune_pathbased}
   \end{subfigure}\hfill
   \begin{subfigure}[t]{.24\textwidth}
     \centering
     \includegraphics[width=\linewidth, height=2.3 cm]{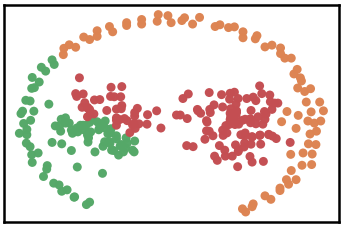}
      \vspace*{-1.0cm}
     \caption{NPIR}\label{Fig:npir_pathbased}
   \end{subfigure}\hfill
   \begin{subfigure}[t]{.24\textwidth}
     \centering
     \includegraphics[width=\linewidth, height=2.3 cm]{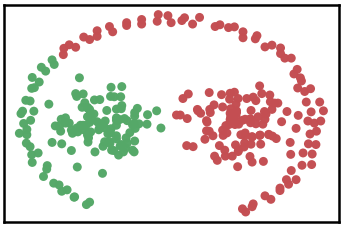}
      \vspace*{-1.0cm}
     \caption{CBKM}\label{Fig:cbkm_pathbased}
   \end{subfigure}\hfill
  
   \begin{subfigure}[t]{.24\textwidth}
    \centering
     \includegraphics[width=\linewidth, height=2.3 cm]{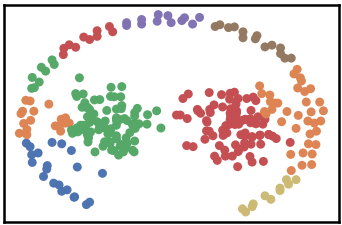}
      \vspace*{-1.0cm}
     \caption{FINCH}\label{Fig:finch_pathbased}
     \end{subfigure}\hfill
   \begin{subfigure}[t]{.24\textwidth}
     \centering
     \includegraphics[width=\linewidth, height=2.3 cm]{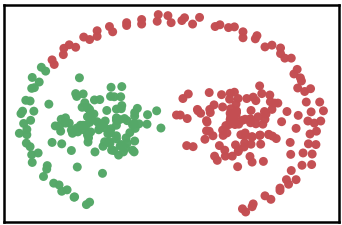}
      \vspace*{-1.0cm}
     \caption{FastDP}\label{Fig:fastdp_pathbased}
   \end{subfigure}\hfill
   \begin{subfigure}[t]{.24\textwidth}
     \centering
     \includegraphics[width=\linewidth, height=2.3 cm]{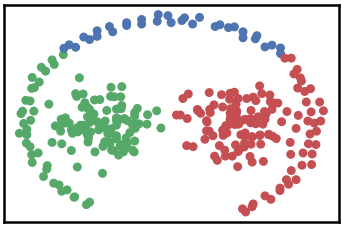}
      \vspace*{-1.0cm}
     \caption{RS}\label{Fig:rs_pathbased}
   \end{subfigure}\hfill
   \begin{subfigure}[t]{.24\textwidth}
     \centering
     \includegraphics[width=\linewidth, height=2.3 cm]{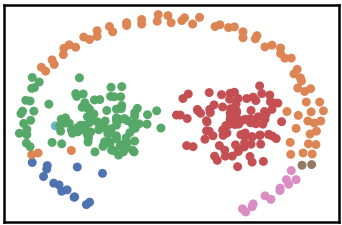}
      \vspace*{-1.0cm}
     \caption{RCC}\label{Fig:rcc_pathbased}
   \end{subfigure}\hfill

  \begin{subfigure}[t]{.24\textwidth}
    \centering
     \includegraphics[width=\linewidth, height=2.3 cm]{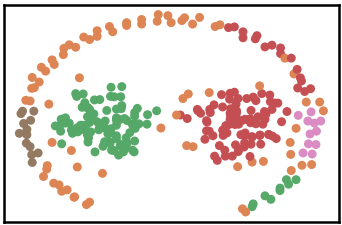}
     \vspace*{-1.0cm}
     \caption{HDBSCAN}\label{Fig:hdbscan_pathbased}
     \end{subfigure}\hfill
   \begin{subfigure}[t]{.24\textwidth}
     \centering
     \includegraphics[width=\linewidth, height=2.3 cm]{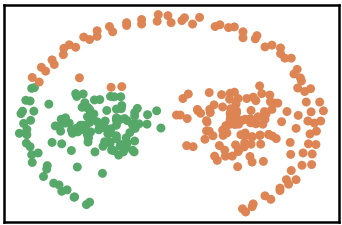}
      \vspace*{-1.0cm}
     \caption{Spectral clustering}\label{Fig:spectral_pathbased}
   \end{subfigure}\hfill
   \begin{subfigure}[t]{.24\textwidth}
     \centering
     \includegraphics[width=\linewidth, height=2.3 cm]{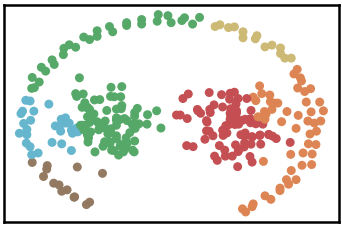}
      \vspace*{-1.0cm}
     \caption{KMeans++}\label{Fig:kmeanspp_pathbased}
   \end{subfigure}\hfill
   \begin{subfigure}[t]{.24\textwidth}
     \centering
     \includegraphics[width=\linewidth, height=2.3 cm]{images/visualize/empty.png}
      \vspace*{-1.0cm}
    
   \end{subfigure}\hfill

   \end{minipage}\hfill
    \vspace*{-0.3cm}
   \caption{Visualization of the results obtained by the ten algorithms for the  Pathbased dataset.}
   \label{Fig:Pathbased_Results}

   \end{figure}

On contrary to DenMune, density-based algorithms such as HDBSCAN fails when clusters have different densities, that is why  HDBSCAN performs badly on Compound and Pathbased datsets, Fig. \ref{Fig:hdbscan_compound} and Fig. \ref{Fig:hdbscan_pathbased}, respectively. Also, on Aggregation dataset it merges some spherical shapes incorrectly due to the strong linkage between them, Fig. \ref{Fig:hdbscan_aggregation}. nevertheless, it performs well on spiral dataset, Fig.\ref{Fig:hdbscan_spiral} since clusters are well separated.

FastDP speeds up the clustering process by building an approximate k-nearest neighbor (kNN) graph using an iterative algorithm. Its main advantage is that it removes the quadratic time complexity limitation of density peaks and allows clustering of very large datasets. FastDP can not select the right cluster centers on Pathbased and Spiral  datasets Fig. (\ref{Fig:fastdp_pathbased} and \ref{Fig:fastdp_spiral}) respectively. Its performance goes down when working on datasets with extremely uneven distributions as in Compound dataset, Fig. \ref{Fig:fastdp_compound}. On contrary to the speed achieved by the algorithm, results obtained showed lower clustering quality. This is obvious from the validity indexes values achieved by the algorithm.

Centroid based algorithms fail when the centroid of a cluster is closer to other data points rather than the data points of its representative cluster, That is why KMeans++ and spectral clustering perform badly on arbitrary shaped data.  They can detect clusters of globular shapes specifically when clusters are well separated as in DIM datasets. Datasets with varying clusters' overlap degrade validations scores even if the clusters are of globular shapes as in G2 datasets.  Although,they have an advantage over traditional KMmeans, they perform badly on  noisy or data with overlapping clusters.

\begin{figure}[H]
   
  \begin{minipage}[b]{\linewidth}
   \centering
   
      \begin{subfigure}[t]{.24\textwidth}
    \centering
     \includegraphics[width=\linewidth, height=2.3 cm]{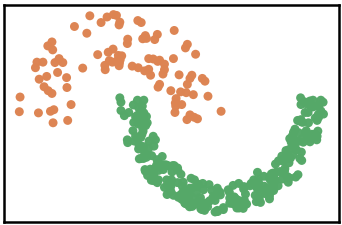}
      \vspace*{-1.0cm}
     \caption{Jain dataset}\label{Fig:ground_jain}
     \end{subfigure}\hfill
   \begin{subfigure}[t]{.24\textwidth}
     \centering
     \includegraphics[width=\linewidth, height=2.3 cm]{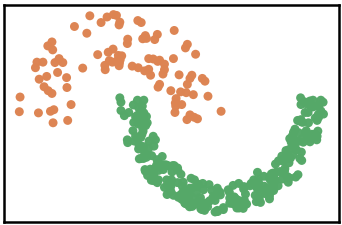}
      \vspace*{-1.0cm}
     \caption{DenMune}\label{Fig:denmune_jain}
   \end{subfigure}\hfill
   \begin{subfigure}[t]{.24\textwidth}
     \centering
     \includegraphics[width=\linewidth, height=2.3 cm]{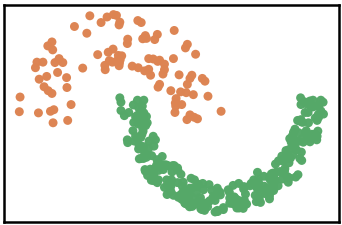}
      \vspace*{-1.0cm}
     \caption{NPIR}\label{Fig:npir_jain}
   \end{subfigure}\hfill
   \begin{subfigure}[t]{.24\textwidth}
     \centering
     \includegraphics[width=\linewidth, height=2.3 cm]{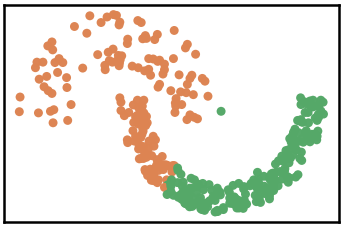}
      \vspace*{-1.0cm}
     \caption{CBKM}\label{Fig:cbkm_jain}
   \end{subfigure}\hfill
  
   \begin{subfigure}[t]{.24\textwidth}
    \centering
     \includegraphics[width=\linewidth, height=2.3 cm]{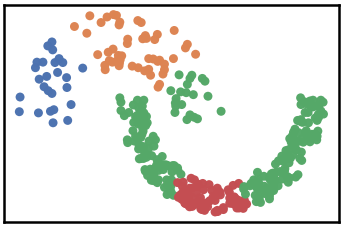}
      \vspace*{-1.0cm}
     \caption{FINCH}\label{Fig:finch_jain}
     \end{subfigure}\hfill
   \begin{subfigure}[t]{.24\textwidth}
     \centering
     \includegraphics[width=\linewidth, height=2.3 cm]{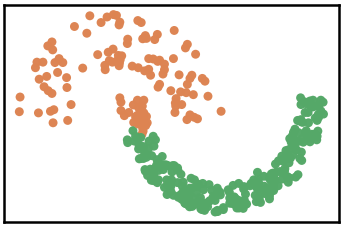}
      \vspace*{-1.0cm}
     \caption{FastDP}\label{Fig:fastdp_jain}
   \end{subfigure}\hfill
   \begin{subfigure}[t]{.24\textwidth}
     \centering
     \includegraphics[width=\linewidth, height=2.3 cm]{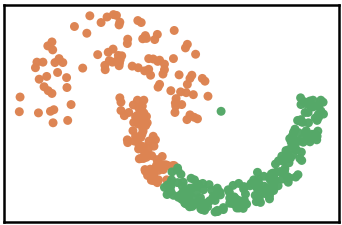}
      \vspace*{-1.0cm}
     \caption{RS}\label{Fig:rs_jain}
   \end{subfigure}\hfill
   \begin{subfigure}[t]{.24\textwidth}
     \centering
     \includegraphics[width=\linewidth, height=2.3 cm]{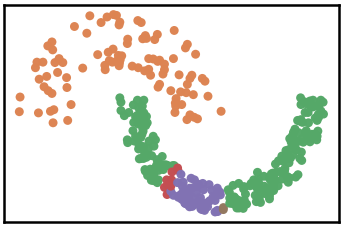}
      \vspace*{-1.0cm}
     \caption{RCC}\label{Fig:rcc_jain}
   \end{subfigure}\hfill

  \begin{subfigure}[t]{.24\textwidth}
    \centering
     \includegraphics[width=\linewidth, height=2.3 cm]{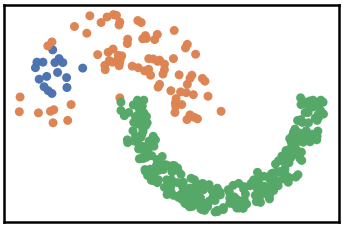}
     \vspace*{-1.0cm}
     \caption{HDBSCAN}\label{Fig:hdbscan_jain}
     \end{subfigure}\hfill
   \begin{subfigure}[t]{.24\textwidth}
     \centering
     \includegraphics[width=\linewidth, height=2.3 cm]{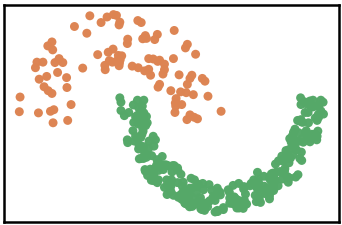}
      \vspace*{-1.0cm}
     \caption{Spectral clustering}\label{Fig:spectral_jain}
   \end{subfigure}\hfill
   \begin{subfigure}[t]{.24\textwidth}
     \centering
     \includegraphics[width=\linewidth, height=2.3 cm]{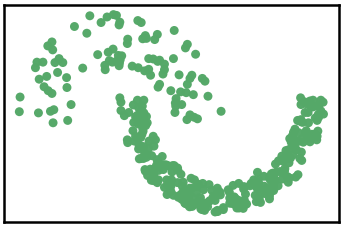}
      \vspace*{-1.0cm}
     \caption{KMeans++}\label{Fig:kmeanspp_jain}
   \end{subfigure}\hfill
   \begin{subfigure}[t]{.24\textwidth}
     \centering
     \includegraphics[width=\linewidth, height=2.3 cm]{images/visualize/empty.png}
      \vspace*{-1.0cm}
    
   \end{subfigure}\hfill

   \end{minipage}\hfill
    \vspace*{-0.3cm}
   \caption{Visualization of the results obtained by the ten algorithms for the Jain dataset.}
   \label{Fig:jain_Results}

   \end{figure}

NPIR uses an indexing ratio, IR to control the amount of possible reassignment of points. The higher IR value means that the assigned points have more possibility for reassignment. The reassignment process does not guarantee algorithm to assign points to the correct clusters specifically when data are noisy as in A1 and A2 datasets (F1= 0.48 and 0.40 respectively) or clusters with different degree of cluster overlap as in S1 and S1 datasets (F1= 0.43 and 0.41 respectively). It is obvious that NPIR performs badly when data are noisy or clusters are of different densities and attached to each other. It performs better when clusters are well separated even if they are of different densities as in Jain and Aggregation datasets  Fig.\ref{Fig:npir_jain} and Fig.\ref{Fig:npir_aggregation}, respectively. A noticeable issue we experienced during examining the algorithm is that it could not scale when working on the MNIST dataset and failed to run even on a cloud server with 128 GB memory. We tested NPIR for IR in the range [0.01, 0.05, 0.10, 0.15, 0.20.]. We found that NPIR performs well when IR is set to 0.01. However, it achieved its highest score for some datasets such as Mouse, Ecoli and Compound datasets for IR=0.15 and for Flame dataset on IR=0.20. Tuning NPIR to yield the best results is not an easy task.

We can observe easily that the performance of all clustering algorithms decrease for datasets with clusters' overlap as in S1, S2, A1 and A2 datasets except for DenMune algorithm. DenMune can deal with overlaping clusters as long as they are of distinguishable densities. DenMune outperformed the other algorithms, for these datasets, with a remarkable margin.  

RCC performs well on some datasets such as G2, Spiral and R15 datatsets, but it performs too badly on DIM datasets, 	a high-dimensional datasets where clusters are well separated even in the higher dimensional space, Fig(\ref{Fig:ground_dim032}:\ref{Fig:ground_dim512}) .

\begin{figure}[H]
   
  \begin{minipage}[b]{\linewidth}
   \centering
   
      \begin{subfigure}[t]{.24\textwidth}
    \centering
     \includegraphics[width=\linewidth, height=2.3 cm]{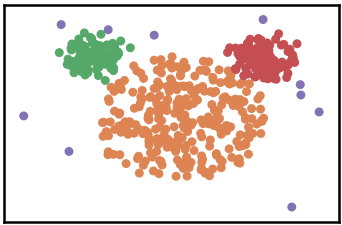}
      \vspace*{-1.0cm}
     \caption{Mouse dataset}\label{Fig:ground_mouse}
     \end{subfigure}\hfill
   \begin{subfigure}[t]{.24\textwidth}
     \centering
     \includegraphics[width=\linewidth, height=2.3 cm]{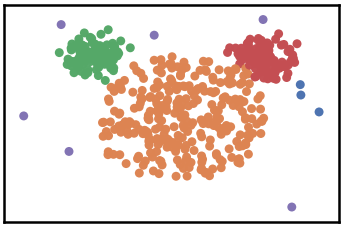}
      \vspace*{-1.0cm}
     \caption{DenMune}\label{Fig:denmune_mouse}
   \end{subfigure}\hfill
   \begin{subfigure}[t]{.24\textwidth}
     \centering
     \includegraphics[width=\linewidth, height=2.3 cm]{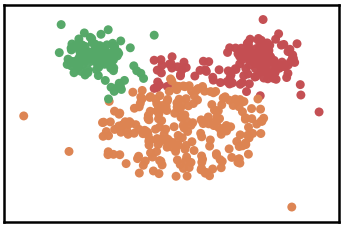}
      \vspace*{-1.0cm}
     \caption{NPIR}\label{Fig:npir_mouse}
   \end{subfigure}\hfill
   \begin{subfigure}[t]{.24\textwidth}
     \centering
     \includegraphics[width=\linewidth, height=2.3 cm]{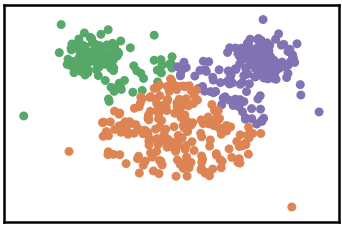}
      \vspace*{-1.0cm}
     \caption{CBKM}\label{Fig:cbkm_mouse}
   \end{subfigure}\hfill
  
   \begin{subfigure}[t]{.24\textwidth}
    \centering
     \includegraphics[width=\linewidth, height=2.3 cm]{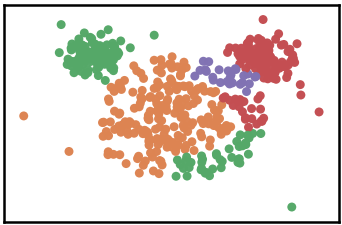}
      \vspace*{-1.0cm}
     \caption{FINCH}\label{Fig:finch_mouse}
     \end{subfigure}\hfill
   \begin{subfigure}[t]{.24\textwidth}
     \centering
     \includegraphics[width=\linewidth, height=2.3 cm]{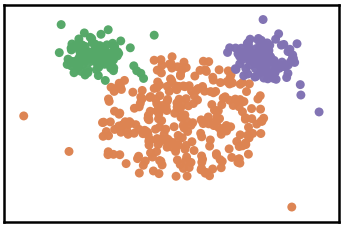}
      \vspace*{-1.0cm}
     \caption{FastDP}\label{Fig:fastdp_mouse}
   \end{subfigure}\hfill
   \begin{subfigure}[t]{.24\textwidth}
     \centering
     \includegraphics[width=\linewidth, height=2.3 cm]{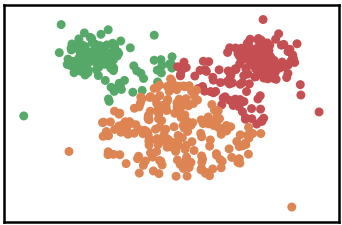}
      \vspace*{-1.0cm}
     \caption{RS}\label{Fig:rs_mouse}
   \end{subfigure}\hfill
   \begin{subfigure}[t]{.24\textwidth}
     \centering
     \includegraphics[width=\linewidth, height=2.3 cm]{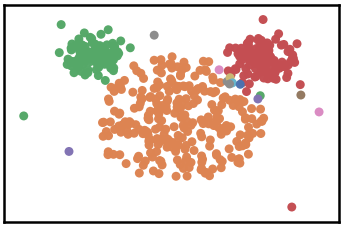}
      \vspace*{-1.0cm}
     \caption{RCC}\label{Fig:rcc_mouse}
   \end{subfigure}\hfill

  \begin{subfigure}[t]{.24\textwidth}
    \centering
     \includegraphics[width=\linewidth, height=2.3 cm]{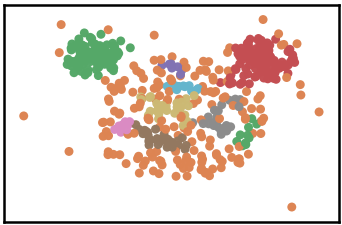}
     \vspace*{-1.0cm}
     \caption{HDBSCAN}\label{Fig:hdbscan_mouse}
     \end{subfigure}\hfill
   \begin{subfigure}[t]{.24\textwidth}
     \centering
     \includegraphics[width=\linewidth, height=2.3 cm]{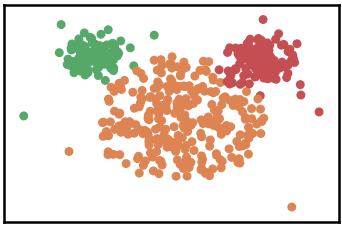}
      \vspace*{-1.0cm}
     \caption{Spectral clustering}\label{Fig:spectral_mouse}
   \end{subfigure}\hfill
   \begin{subfigure}[t]{.24\textwidth}
     \centering
     \includegraphics[width=\linewidth, height=2.3 cm]{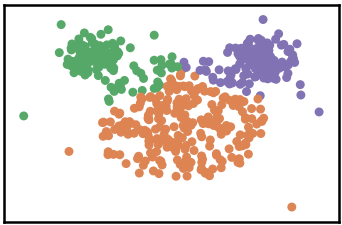}
      \vspace*{-1.0cm}
     \caption{KMeans++}\label{Fig:kmeanspp_mouse}
   \end{subfigure}\hfill
   \begin{subfigure}[t]{.24\textwidth}
     \centering
     \includegraphics[width=\linewidth, height=2.3 cm]{images/visualize/empty.png}
      \vspace*{-1.0cm}
    
   \end{subfigure}\hfill

   \end{minipage}\hfill
    \vspace*{-0.3cm}
   \caption{Visualization of the results obtained by the ten algorithms for the Mouse dataset.}
   \label{Fig:Mouse_Results}

   \end{figure}

For synthetic datasets (\ref{Tab:f1_synthetic}),  FINCH has the closest F1-score to DenMune while being faster. However, for the same datasets, its performance based on the NMI and ARI metrics (Tables  \ref{Tab:nmi-real} and \ref{Tab:ari-real} as well as  Figs \ref{Fig:finch_aggregation} to  \ref{Fig:finch_mouse} ) is bad. The same applies for real datasets ( \ref{Tab:f1_real}).

RS algorithm adopts a randomized search strategy, which is simple to implement and efficient. It archived good quality clustering, and if iterated longer, it would finds the correct clustering with high probability. CBKM algorithm uses a better initialization technique and/or  repeating (restarting) the algorithm to improve the quality clustering of KMeans to overcome issues with clusters overlap and clusters of unbalanced sizes. Authors of CBKM  observed that choosing an initialization technique like Maxmin can compensate for the weaknesses of k-means and recommended  that repeating k-means 10–100 times; each time taking a random point as the first centroids and selecting the rest using the Maxmin heuristic would improve the quality clustering. We believe that increasing the number of runs (from 100 to say 1000) would slightly increase the clustering quality of RS, CBKM and NBIR since they have random initial states. We found that RS and CBKM algorithms have the most reasonable results achieved for both real and synthetic datasets assessed by F1, NMI and ARI scores. In general, they have the closest rank to DenMune.

Also, we found that DenMune performs moderately for small size datasets where DenMune has not enough chance to build robust $KNN$ framework to distinguish clusters, this is the case for Iris and Arcene datasets.

Finally, we recorded the Homogeneity and Completeness of DenMune in Tables(\ref{Tab:denmune_homogeneity} and \ref{Tab:denmune_completeness}). A clustering result satisfies homogeneity if each cluster contains only members of a single class, while it satisfies completeness if all members of a given class are assigned to the same cluster. It is easy to observe that DenMune has high homogeneity and completeness scores, which explain the goodness of its clustering quality.

\begin{table}[!htb]
\centering
\caption{Best NMI scores, obtained by DenMune, when applying   different dimensionality reduction methods on three real N-D datasets}
\label{tab:reduction_alg}
\begin{tabular}{lllll}

\hline
Dataset & PCA    &  FA & NMF  & t-sne       \\
\hline
\hline

Optical Digits  & 0.43            & 0.46 & 0.31  & 0.95 \\
Pen Digits  & 0.48        & 0.46 & 0.36  & 0.88 \\
MNIST  & 0.25    & 0.25 & 0.24  & 0.89 \\
\hline

\end{tabular}
\end{table}

\begin{table}[H]
\caption{Best NMI scores when applying DenMune on five real datasets before and after dimensionality reduction}
\label{tab:2d_nd}
\centering
\begin{tabular}{lcccc}
\hline
\multicolumn{1}{l|}{\multirow{2}{*}{Dataset}} & \multicolumn{2}{c|}{Original dataset}                      & \multicolumn{2}{c}{Reduced dataset}                       \\ \cline{2-5} 
\multicolumn{1}{l|}{}                         & \multicolumn{1}{c|}{Dimensions} & \multicolumn{1}{c|}{NMI} & \multicolumn{1}{c|}{Dimensions} & \multicolumn{1}{c}{NMI} \\ \hline \hline
Optical Digits & 64    & 0.75 & 2 & 0.95 \\
SCC            & 60    & 0.85 & 2 & 0.85 \\
Arcene         & 10000 & 0.19 & 2 & 0.20 \\
iris           & 4     & 0.73 & 2 & 0.81 \\
Breast Cancer  & 9     & 0.75 & 2 & 0.80 \\
Ecoli          & 8     & 0.01 & 2 & 0.71 \\
Pen Digits     & 16    & 0.81 & 2 & 0.88 \\
\hline
\end{tabular}%
\end{table}

\subsection{Speed Performance}
\begin{figure}[H]
\begin{minipage}[b]{\linewidth}
%
\begin{subfigure}[t]{0.98\textwidth}
 	\raisebox{-\height}{\includegraphics[width=\textwidth, height=5.0 cm]{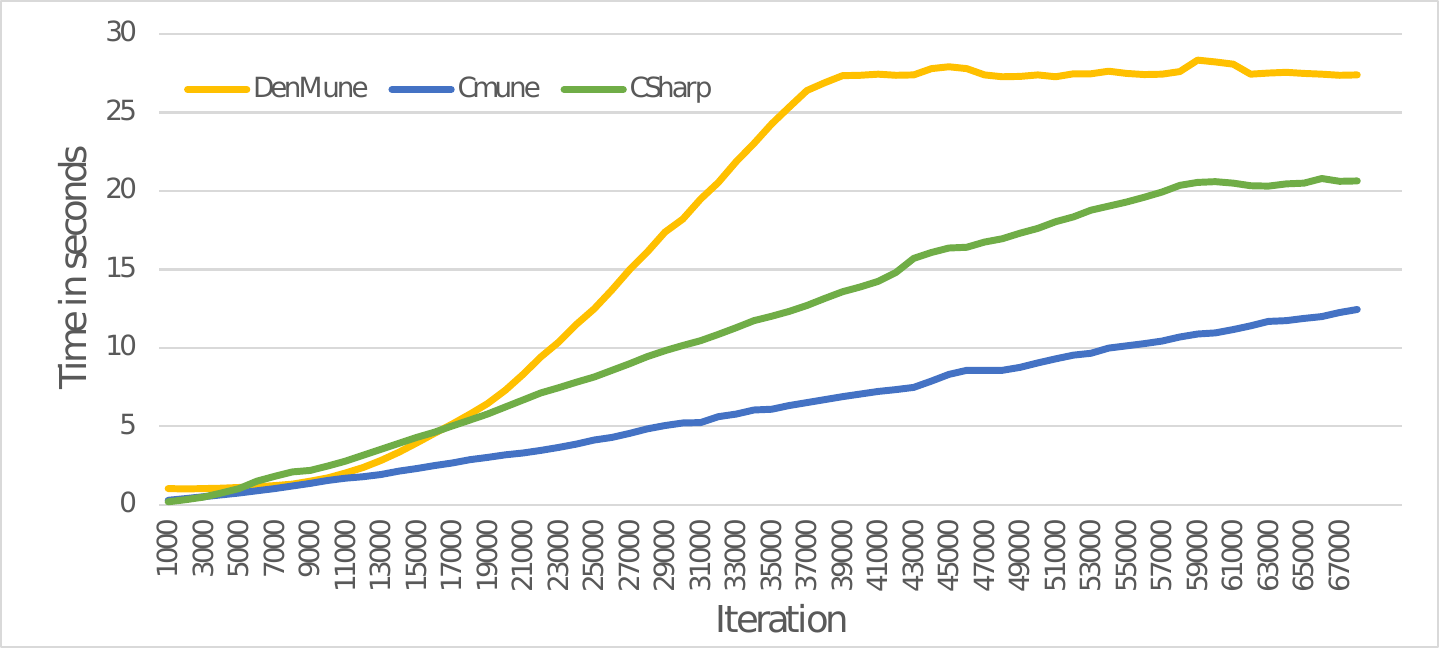}}
      \caption{DenMune speed performance}
       
\end{subfigure}
  \hfill
\begin{subfigure}[t]{0.98\textwidth}
	\raisebox{-\height}{\includegraphics[width=\textwidth, height=5.0 cm]{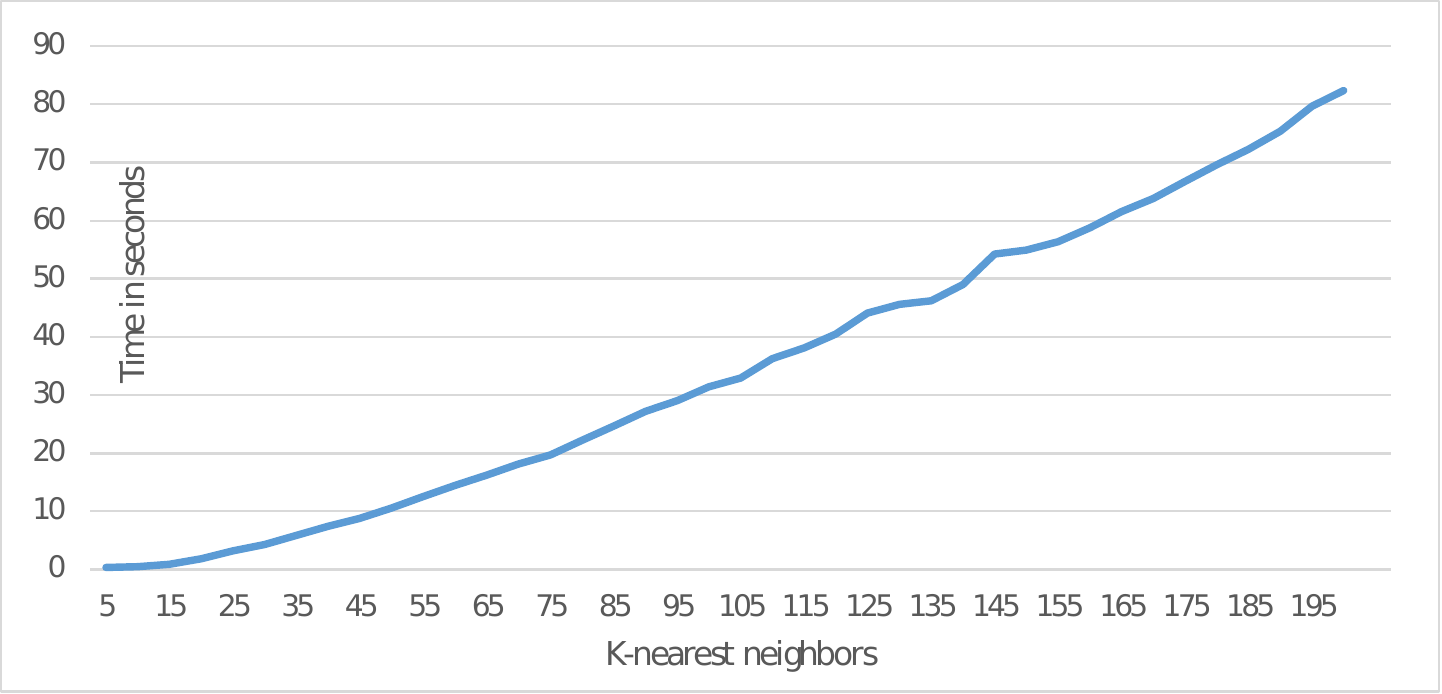}}
         \caption{DenMune : Speed vs number of $K$- nearest neighbors. }
    \end{subfigure}
\hfill
\end{minipage}%
\caption{ (a) Speed of DenMune compared to the speed of CMune and CSharp on the MNIST dataset. (b) Speed of DenMune vs number of $K$- nearest neighbors.}
\label{fig:spped-comparison}
\end{figure}

The speed of DenMune has been compared to the speed of CMune and CSharp, as shown in Fig.\ref{fig:spped-comparison}a. The data set
considered is the MNIST dataset (with 70000 patterns), after dividing it into subsets, each of size 1000 patterns. The subsets are added incrementally, and the speed of the
algorithm is recorded with each increment. The time considered is the time required for running the core clustering algorithms, excluding the pre-processing time for computing the proximity matrix and dimensionality reduction. The time is measured in seconds. The adopted algorithms as well as the proposed algorithm have been executed on a cloud with the following configuration: Intel E5 Processor, up to 128 GB RAM, and running Linux operating system (Ubuntu 18.04 LTS).
Another test is conducted to examine speed versus the number of  K-nearest neighbors used, as shown in Fig.\ref{fig:spped-comparison}b

\section{Conclusion and Future Work}

\begin{figure}[H]
\centering
\includegraphics[height=6.0 cm] {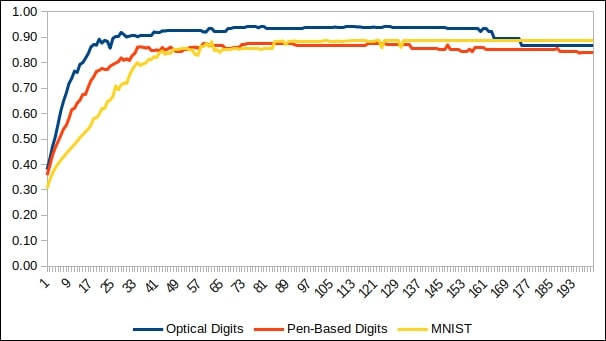}
\caption {DenMune Results stability over changes in K, measured in NMI}
\label{Fig:denmue_stability}      
\end{figure}

In this paper, a novel shared nearest neighbors clustering algorithm DenMune, is presented. It utilizes the MNN size to calculate the density of each point and chooses the high-density points as the seeds from which clusters may grow up. In contrast to recent similar algorithms, such as DPC and CMune, no cut-off parameter is needed from the user of DenMune. Guided by the principle of Mutual Nearest-Neighbors (\textit{MNN}) consistency, DenMune  prioritizes points according to a voting system and partitions them into seeds and non-seeds. Seed points determine the number as well as the skeleton of the  clusters  while non-seed points either merge with the formed clusters or are considered as noise. It has the ability to automatically detect the number of clusters and has shown robustness for datasets of different shapes and densities. We examined the sensitivity of DenMune to changes in K, the number of nearest neighbors (the only parameter required by the algorithm) on three real datasets with $K$ in the range [1..200] and recorded the NMI for each dataset as shown in Fig. \ref{Fig:denmue_stability}. The stability of DenMune, with respect to K, makes it a good candidate for data exploration and visualization since it works in a two-dimensional feature space. Algorithms that rely on several parameters  such as CSharp, CMune, HDBSCAN and  DPC can offer more flexibility than single parameter algorithms such as DenMune, but at the expense of the time needed for their tuning. 

Although the motivations behind the algorithm are logical (the scheme adopted by the algorithm to partition points in a given data set into three  types (seed, noise and potential noise points) and the MNN consistency principle that governs clusters growth), the conducted experiments on a variety of data sets, have shown its efficiency and robustness in detecting clusters of different sizes, shapes and densities in the presence of noise. In summary, DenMune is conceptually simple, logically sound, relies on a single parameter. 
As future work, we intend to implement a parallel version of it, since clusters' propagation in the algorithm  is inherently parallel, as shown in Fig.\ref{Fig:DenMune-Propagation}, and investigate its performance on other types of datasets. 

\section*{Acknowledgement}
The authors would like to thank the anonymous reviewers for their valuable comments and suggestions which contributed to the improvement of the manuscript. Also, the authors would like to thank Dr. Saquib Sarfraz, Dr. Sami Sieranoja and Dr. Pasi Fränti, the authors of FINCH, CBKM and RS algorithms, respectively.


\begingroup
\setlength{\tabcolsep}{3pt} 
\renewcommand{\arraystretch}{1.2} 
\centering
\begin{table}[H]
\caption{Comparison of the performance of DenMune with other nine algorithms, based on F1-score, on twenty-one synthetic datasets.}
\label{Tab:f1_synthetic}
\centering
\resizebox{\textwidth}{!}{%
\begin{tabular}{lcccccccccc}
\hline
{\color[HTML]{FE0000} Dataset} &
  {\color[HTML]{FE0000} DenMune} &
  {\color[HTML]{FE0000} NPIR} &
  {\color[HTML]{FE0000} CBKM} &
  {\color[HTML]{FE0000} RS} &
  {\color[HTML]{FE0000} FastDP} &
  {\color[HTML]{FE0000} FINCH} &
  {\color[HTML]{FE0000} HDBSCAN} &
  {\color[HTML]{FE0000} RCC} &
  {\color[HTML]{FE0000} Spectral} &
  {\color[HTML]{FE0000} KM++} \\
  \hline
  \hline
A1           & 0.93 & 0.48 & 0.55 & 0.60 & 0.44 & 0.80 & 0.40 & 0.57 & 0.48 & 0.63 \\
A2           & 0.95 & 0.40 & 0.45 & 0.41 & 0.48 & 0.81 & 0.38 & 0.61 & 0.46 & 0.57 \\
Aggregation  & 1.00 & 0.65 & 0.77 & 0.73 & 0.70 & 0.69 & 0.66 & 0.81 & 0.67 & 0.69 \\
Compound     & 0.97 & 0.62 & 0.58 & 0.57 & 0.54 & 0.77 & 0.63 & 0.89 & 0.40 & 0.65 \\
D31          & 0.97 & 0.42 & 0.52 & 0.48 & 0.41 & 0.63 & 0.38 & 0.59 & 0.46 & 0.54 \\
Dim-32       & 1.00 & 0.52 & 0.75 & 0.58 & 0.51 & 0.99 & 0.45 & 0.07 & 0.54 & 0.92 \\
Dim-128      & 1.00 & 0.53 & 0.58 & 0.54 & 0.59 & 0.80 & 0.52 & 0.10 & 0.58 & 0.94 \\
Dim-512      & 1.00 & 0.53 & 0.67 & 0.59 & 0.59 & 0.22 & 0.36 & 0.01 & 0.68 & 0.97 \\
Flame        & 1.00 & 1.00 & 0.86 & 0.86 & 1.00 & 0.99 & 0.91 & 0.80 & 0.98 & 0.85 \\
G2-2-10      & 1.00 & 1.00 & 1.00 & 1.00 & 1.00 & 0.90 & 0.84 & 1.00 & 0.33 & 0.86 \\
G2-2-30      & 0.99 & 0.97 & 0.99 & 0.99 & 0.33 & 0.99 & 0.96 & 0.99 & 0.33 & 0.99 \\
G2-2-50      & 0.90 & 0.90 & 0.92 & 0.92 & 0.92 & 0.69 & 0.65 & 0.67 & 0.33 & 0.53 \\
Jain         & 1.00 & 1.00 & 0.80 & 0.80 & 0.93 & 0.77 & 0.97 & 0.89 & 1.00 & 0.79 \\
Mouse        & 0.98 & 0.88 & 0.66 & 0.82 & 0.76 & 0.82 & 0.77 & 0.96 & 0.95 & 0.67 \\
Pathbased    & 0.97 & 0.84 & 0.49 & 0.53 & 0.50 & 0.73 & 0.77 & 0.87 & 0.48 & 0.69 \\
R15          & 1.00 & 0.56 & 0.56 & 0.65 & 0.39 & 0.91 & 0.47 & 0.99 & 0.56 & 0.62 \\
S1           & 1.00 & 0.43 & 0.58 & 0.49 & 0.42 & 0.78 & 0.49 & 0.75 & 0.67 & 0.72 \\
S2           & 0.97 & 0.41 & 0.58 & 0.58 & 0.22 & 0.72 & 0.58 & 0.49 & 0.63 & 0.70 \\
Spiral       & 1.00 & 0.48 & 0.28 & 0.36 & 0.55 & 0.46 & 1.00 & 0.56 & 1.00 & 0.44 \\
Unbalance    & 1.00 & 0.92 & 0.98 & 0.98 & 0.86 & 0.67 & 0.93 & 0.97 & 0.85 & 0.63 \\
Vary density & 1.00 & 1.00 & 0.95 & 0.95 & 0.56 & 0.67 & 0.95 & 0.89 & 1.00 & 0.77 \\
 \hline
Total Rank   & 24   & 120  & 110  & 107  & 139  & 99   & 147  & 102  & 125  & 117  \\
Avg rank     & 1.14 & 5.71 & 5.24 & 5.10 & 6.62 & 4.71 & 7.00 & 4.86 & 6.0  & 5.57 \\
 \hline
  \hline
{\color[HTML]{FE0000} Rank} &
  {\color[HTML]{FE0000} 1} &
  {\color[HTML]{FE0000} 7} &
  {\color[HTML]{FE0000} 5} &
  {\color[HTML]{FE0000} 4} &
  {\color[HTML]{FE0000} 9} &
  {\color[HTML]{FE0000} 2} &
  {\color[HTML]{FE0000} 10} &
  {\color[HTML]{FE0000} 3} &
  {\color[HTML]{FE0000} 8} &
  {\color[HTML]{FE0000} 6}

\end{tabular}%

}
\end{table}

\endgroup

\begingroup
\setlength{\tabcolsep}{3pt} 
\renewcommand{\arraystretch}{1.5} 
\centering

\begin{table}[H]
\caption{Comparison of the performance of DenMune with other nine algorithms, based on F1-score,  on fifteen real datasets.}
\label{Tab:f1_real}
\centering
\resizebox{\textwidth}{!}{%
\begin{tabular}{lcccccccccc}
\hline
{\color[HTML]{FE0000} Dataset} &
  {\color[HTML]{FE0000} DenMune} &
  {\color[HTML]{FE0000} NPIR} &
  {\color[HTML]{FE0000} CBKM} &
  {\color[HTML]{FE0000} RS} &
  {\color[HTML]{FE0000} FastDP} &
  {\color[HTML]{FE0000} FINCH} &
  {\color[HTML]{FE0000} HDBSCAN} &
  {\color[HTML]{FE0000} RCC} &
  {\color[HTML]{FE0000} Spectral} &
  {\color[HTML]{FE0000} KM++} \\
\hline
\hline
Appendicitis     & 0.89 & 0.89 & 0.75 & 0.75 & 0.88 & 0.76 & 0.86 & 0.73 & 0.71 & 0.78 \\
Arcene           & 0.58 & 0.66 & 0.66 & 0.66 & 0.66 & 0.50 & 0.66 & 0.58 & 0.66 & 0.66 \\
Breast cancer    & 0.97 & 0.97 & 0.88 & 0.91 & 0.97 & 0.48 & 0.96 & 0.54 & 0.51 & 0.96 \\
Optical digits   & 0.97 & 0.49 & 0.69 & 0.58 & 0.59 & 0.54 & 0.70 & 0.58 & 0.58 & 0.59 \\
Pendigits        & 0.89 & 0.53 & 0.65 & 0.65 & 0.41 & 0.57 & 0.52 & 0.62 & 0.52 & 0.67 \\
Ecoli            & 0.77 & 0.69 & 0.71 & 0.65 & 0.28 & 0.52 & 0.21 & 0.53 & 0.51 & 0.56 \\
Glass            & 0.57 & 0.51 & 0.55 & 0.54 & 0.36 & 0.52 & 0.47 & 0.52 & 0.59 & 0.46 \\
Iris             & 0.90 & 0.97 & 0.91 & 0.94 & 0.90 & 0.90 & 0.56 & 0.90 & 0.98 & 0.88 \\
MNIST            & 0.90 & N/A$^{*}$ & 0.66 & 0.64 & 0.46 & 0.61 & 0.84 & 0.18 & 0.83 & 0.61 \\
Libras movement  & 0.46 & 0.27 & 0.29 & 0.33 & 0.24 & 0.44 & 0.27 & 0.41 & 0.26 & 0.39 \\
Robot navigation & 0.60 & 0.43 & 0.49 & 0.46 & 0.38 & 0.57 & 0.56 & 0.27 & 0.43 & 0.56 \\
SCC              & 0.68 & 0.65 & 0.65 & 0.84 & 0.49 & 0.64 & 0.36 & 0.77 & 0.64 & 0.55 \\
Seeds            & 0.89 & 0.91 & 0.90 & 0.89 & 0.54 & 0.76 & 0.82 & 0.89 & 0.52 & 0.77 \\
WDBC             & 0.84 & 0.91 & 0.83 & 0.86 & 0.89 & 0.82 & 0.82 & 0.56 & 0.48 & 0.81 \\
Yeast            & 0.40 & 0.35 & 0.44 & 0.40 & 0.27 & 0.41 & 0.31 & 0.45 & 0.40 & 0.42 \\
\hline
Total Rank       & 37   & 60   & 54   & 57   & 92   & 86   & 88   & 80   & 91   & 72   \\
Avg rank         & 2.64 & 4.29 & 3.86 & 4.07 & 6.57 & 6.14 & 6.29 & 5.71 & 6.5  & 5.14 \\
\hline
\hline
{\color[HTML]{FE0000} Rank} &
  {\color[HTML]{FE0000} 1} &
  {\color[HTML]{FE0000} 4} &
  {\color[HTML]{FE0000} 2} &
  {\color[HTML]{FE0000} 3} &
  {\color[HTML]{FE0000} 10} &
  {\color[HTML]{FE0000} 7} &
  {\color[HTML]{FE0000} 8} &
  {\color[HTML]{FE0000} 6} &
  {\color[HTML]{FE0000} 9} &
  {\color[HTML]{FE0000} 5}
\end{tabular}%
}
\end{table}
\endgroup

\begingroup
\noindent
\tiny
* NPIR failed to scale to adapt to MNIST dataset even on a cloud server with 128 GB memory.
\endgroup

\begingroup
\setlength{\tabcolsep}{3pt} 
\renewcommand{\arraystretch}{1.2} 
\centering
\begin{table}[H]
\caption{Comparison of the performance  of DenMune  with other nine algorithms, based on NMI-score, on twenty-one synthetic datasets.}
\label{Tab:nmi-synthetic}
\centering
\resizebox{\textwidth}{!}{%
\begin{tabular}{ccccccccccc}
\hline
{\color[HTML]{FE0000} Dataset} &
  {\color[HTML]{FE0000} DenMune} &
  {\color[HTML]{FE0000} NPIR} &
  {\color[HTML]{FE0000} CBKM} &
  {\color[HTML]{FE0000} RS} &
  {\color[HTML]{FE0000} FastDP} &
  {\color[HTML]{FE0000} FINCH} &
  {\color[HTML]{FE0000} HDBSCAN} &
  {\color[HTML]{FE0000} RCC} &
  {\color[HTML]{FE0000} Spectral} &
  {\color[HTML]{FE0000} KM++} \\
  \hline
  \hline
A1           & 0.98 & 0.84 & 0.87 & 0.92 & 0.83 & 0.85 & 0.74 & 0.78 & 0.88 & 0.89 \\
A2           & 0.98 & 0.86 & 0.89 & 0.88 & 0.89 & 0.89 & 0.75 & 0.80 & 0.90 & 0.85 \\
Aggregation  & 0.99 & 0.80 & 0.90 & 0.90 & 0.87 & 0.76 & 0.86 & 0.82 & 0.90 & 0.73 \\
Compound     & 0.94 & 0.79 & 0.58 & 0.62 & 0.46 & 0.83 & 0.66 & 0.87 & 0.62 & 0.66 \\
D31          & 0.96 & 0.84 & 0.87 & 0.87 & 0.84 & 0.85 & 0.71 & 0.85 & 0.85 & 0.84 \\
Dim-32       & 1.00 & 0.91 & 0.95 & 0.92 & 0.89 & 0.99 & 0.85 & 0.16 & 0.87 & 0.90 \\
Dim-128      & 1.00 & 0.87 & 0.92 & 0.87 & 0.91 & 0.90 & 0.87 & 0.22 & 0.92 & 0.91 \\
Dim-512      & 1.00 & 0.88 & 0.93 & 0.91 & 0.91 & 0.59 & 0.83 & 0.00 & 0.93 & 0.94 \\
Flame        & 1.00 & 1.00 & 0.46 & 0.48 & 1.00 & 0.94 & 0.61 & 0.47 & 0.85 & 0.55 \\
G2-2-10      & 1.00 & 1.00 & 1.00 & 1.00 & 1.00 & 0.75 & 0.53 & 0.99 & 0.00 & 0.80 \\
G2-2-30      & 0.94 & 0.83 & 0.92 & 0.92 & 0.00 & 0.92 & 0.76 & 0.94 & 0.00 & 0.92 \\
G2-2-50      & 0.51 & 0.53 & 0.59 & 0.59 & 0.58 & 0.36 & 0.25 & 0.23 & 0.00 & 0.26 \\
Jain         & 1.00 & 1.00 & 0.37 & 0.37 & 0.64 & 0.47 & 0.88 & 0.70 & 1.00 & 0.50 \\
Mouse        & 0.94 & 0.68 & 0.58 & 0.58 & 0.85 & 0.56 & 0.55 & 0.87 & 0.85 & 0.61 \\
Pathbased    & 0.89 & 0.66 & 0.51 & 0.55 & 0.52 & 0.56 & 0.56 & 0.70 & 0.50 & 0.58 \\
R15          & 0.99 & 0.86 & 0.89 & 0.91 & 0.84 & 0.90 & 0.85 & 0.99 & 0.90 & 0.79 \\
S1           & 0.99 & 0.84 & 0.89 & 0.87 & 0.84 & 0.80 & 0.83 & 0.86 & 0.91 & 0.88 \\
S2           & 0.94 & 0.79 & 0.83 & 0.82 & 0.62 & 0.75 & 0.74 & 0.80 & 0.87 & 0.77 \\
Spiral       & 1.00 & 0.28 & 0.00 & 0.00 & 0.74 & 0.26 & 1.00 & 0.50 & 1.00 & 0.31 \\
Unbalance    & 1.00 & 0.94 & 0.99 & 0.99 & 0.82 & 0.97 & 0.95 & 0.95 & 0.88 & 0.90 \\
Vary density & 1.00 & 1.00 & 0.86 & 0.86 & 0.73 & 0.53 & 0.87 & 0.70 & 1.00 & 0.56 \\
\hline
Total Rank   & 25   & 109  & 97   & 99   & 127  & 129  & 151  & 125  & 99   & 130  \\
Avg rank     & 1.19 & 5.19 & 4.62 & 4.71 & 6.05 & 6.14 & 7.19 & 5.95 & 4.7  & 6.19 \\
\hline
\hline
{\color[HTML]{FE0000} Rank} &
  {\color[HTML]{FE0000} 1} &
  {\color[HTML]{FE0000} 5} &
  {\color[HTML]{FE0000} 2} &
  {\color[HTML]{FE0000} 3} &
  {\color[HTML]{FE0000} 7} &
  {\color[HTML]{FE0000} 8} &
  {\color[HTML]{FE0000} 10} &
  {\color[HTML]{FE0000} 6} &
  {\color[HTML]{FE0000} 3} &
  {\color[HTML]{FE0000} 9}
\end{tabular}%
}
\end{table}
\endgroup

\begingroup
\setlength{\tabcolsep}{3pt} 
\renewcommand{\arraystretch}{1.5} 
\centering
\begin{table}[H]
\caption{Comparison of the performance of DenMune with other nine algorithms, based on NMI-score, on fifteen real datasets.}
\label{Tab:nmi-real}
\centering
\resizebox{\textwidth}{!}{%
\begin{tabular}{lcccccccccc}
\hline
{\color[HTML]{FE0000} Dataset} &
  {\color[HTML]{FE0000} DenMune} &
  {\color[HTML]{FE0000} NPIR} &
  {\color[HTML]{FE0000} CBKM} &
  {\color[HTML]{FE0000} RS} &
  {\color[HTML]{FE0000} FastDP} &
  {\color[HTML]{FE0000} FINCH} &
  {\color[HTML]{FE0000} HDBSCAN} &
  {\color[HTML]{FE0000} RCC} &
  {\color[HTML]{FE0000} Spectral} &
  {\color[HTML]{FE0000} KM++} \\
  \hline
  \hline
Appendicitis     & 0.37 & 0.37 & 0.18 & 0.18 & 0.33 & 0.18 & 0.26 & 0.21 & 0.00 & 0.16 \\
Arcene           & 0.07 & 0.09 & 0.09 & 0.09 & 0.09 & 0.07 & 0.09 & 0.02 & 0.09 & 0.09 \\
Breast cancer    & 0.80 & 0.80 & 0.53 & 0.61 & 0.80 & 0.20 & 0.77 & 0.24 & 0.00 & 0.77 \\
Optical digits   & 0.94 & 0.83 & 0.86 & 0.81 & 0.82 & 0.73 & 0.83 & 0.82 & 0.80 & 0.81 \\
Pendigits        & 0.88 & 0.72 & 0.78 & 0.71 & 0.57 & 0.76 & 0.77 & 0.74 & 0.70 & 0.75 \\
Ecoli            & 0.71 & 0.56 & 0.53 & 0.54 & 0.52 & 0.49 & 0.51 & 0.43 & 0.49 & 0.55 \\
Glass            & 0.39 & 0.34 & 0.37 & 0.35 & 0.31 & 0.37 & 0.35 & 0.36 & 0.39 & 0.34 \\
Iris             & 0.81 & 0.90 & 0.79 & 0.82 & 0.81 & 0.81 & 0.73 & 0.80 & 0.92 & 0.82 \\
MNIST            & 0.86 & N/A$^{*}$  & 0.74 & 0.76 & 0.61 & 0.75 & 0.87 & 0.54 & 0.85 & 0.66 \\
Libras movement  & 0.67 & 0.52 & 0.56 & 0.52 & 0.52 & 0.63 & 0.57 & 0.63 & 0.54 & 0.63 \\
Robot navigation & 0.43 & 0.24 & 0.24 & 0.23 & 0.06 & 0.37 & 0.33 & 0.37 & 0.17 & 0.43 \\
SCC              & 0.82 & 0.78 & 0.68 & 0.79 & 0.66 & 0.71 & 0.72 & 0.76 & 0.72 & 0.66 \\
Seeds            & 0.69 & 0.71 & 0.73 & 0.71 & 0.56 & 0.60 & 0.62 & 0.68 & 0.53 & 0.60 \\
WDBC             & 0.46 & 0.56 & 0.41 & 0.45 & 0.50 & 0.47 & 0.41 & 0.32 & 0.00 & 0.47 \\
Yeast            & 0.27 & 0.22 & 0.26 & 0.25 & 0.25 & 0.23 & 0.22 & 0.28 & 0.19 & 0.18 \\
\hline
Total Rank       & 33   & 52   & 66   & 71   & 83   & 81   & 73   & 81   & 99   & 72   \\
Avg rank         & 2.36 & 3.71 & 4.71 & 5.07 & 5.93 & 5.79 & 5.21 & 5.79 & 7.1  & 5.14 \\
\hline
\hline
{\color[HTML]{FE0000} Rank} &
  {\color[HTML]{FE0000} 1} &
  {\color[HTML]{FE0000} 2} &
  {\color[HTML]{FE0000} 3} &
  {\color[HTML]{FE0000} 4} &
  {\color[HTML]{FE0000} 9} &
  {\color[HTML]{FE0000} 7} &
  {\color[HTML]{FE0000} 6} &
  {\color[HTML]{FE0000} 7} &
  {\color[HTML]{FE0000} 10} &
  {\color[HTML]{FE0000} 5}
\end{tabular}%
}
\end{table}

\endgroup

\begingroup
\noindent
\tiny
* NPIR failed to scale to adapt to MNIST dataset even on a cloud server with 128 GB memory.
\endgroup

\begingroup
\setlength{\tabcolsep}{3pt} 
\renewcommand{\arraystretch}{1.2} 
\centering

\begin{table}[H]
\caption{Comparison of the performance of DenMune with other nine algorithms, based on ARI-score, on twenty-one synthetic datasets.}
\label{Tab:ari-synthetic}

\centering
\resizebox{\textwidth}{!}{%
\begin{tabular}{lcccccccccc}
\hline
{\color[HTML]{FE0000} Dataset} &
  {\color[HTML]{FE0000} DenMune} &
  {\color[HTML]{FE0000} NPIR} &
  {\color[HTML]{FE0000} CBKM} &
  {\color[HTML]{FE0000} RS} &
  {\color[HTML]{FE0000} FastDP} &
  {\color[HTML]{FE0000} FINCH} &
  {\color[HTML]{FE0000} HDBSCAN} &
  {\color[HTML]{FE0000} RCC} &
  {\color[HTML]{FE0000} Spectral} &
  {\color[HTML]{FE0000} KM++} \\
  \hline
  \hline
A1           & 0.94 & 0.56 & 0.62 & 0.75 & 0.51 & 0.69 & 0.42 & 0.53 & 0.64 & 0.70 \\
A2           & 0.96 & 0.55 & 0.62 & 0.61 & 0.61 & 0.73 & 0.37 & 0.53 & 0.64 & 0.60 \\
Aggregation  & 0.99 & 0.65 & 0.90 & 0.90 & 0.81 & 0.57 & 0.77 & 0.65 & 0.84 & 0.54 \\
Compound     & 0.97 & 0.73 & 0.56 & 0.59 & 0.30 & 0.79 & 0.62 & 0.86 & 0.36 & 0.44 \\
D31          & 0.94 & 0.55 & 0.64 & 0.65 & 0.56 & 0.63 & 0.33 & 0.65 & 0.55 & 0.60 \\
Dim-32       & 1.00 & 0.73 & 0.83 & 0.74 & 0.67 & 0.97 & 0.55 & 0.02 & 0.58 & 0.84 \\
Dim-128      & 1.00 & 0.57 & 0.74 & 0.58 & 0.70 & 0.76 & 0.62 & 0.03 & 0.74 & 0.89 \\
Dim-512      & 1.00 & 0.60 & 0.78 & 0.70 & 0.70 & 0.16 & 0.54 & 0.00 & 0.74 & 0.93 \\
Flame        & 1.00 & 1.00 & 0.50 & 0.51 & 1.00 & 0.97 & 0.69 & 0.40 & 0.92 & 0.52 \\
G2-2-10      & 1.00 & 1.00 & 1.00 & 1.00 & 1.00 & 0.72 & 0.63 & 1.00 & 0.00 & 0.75 \\
G2-2-30      & 0.97 & 0.89 & 0.96 & 0.96 & 0.00 & 0.96 & 0.84 & 0.97 & 0.00 & 0.96 \\
G2-2-50      & 0.63 & 0.64 & 0.70 & 0.70 & 0.69 & 0.30 & 0.21 & 0.17 & 0.00 & 0.17 \\
Jain         & 1.00 & 1.00 & 0.32 & 0.32 & 0.71 & 0.38 & 0.95 & 0.59 & 1.00 & 0.44 \\
Mouse        & 0.97 & 0.69 & 0.54 & 0.54 & 0.92 & 0.53 & 0.40 & 0.92 & 0.90 & 0.37 \\
Pathbased    & 0.92 & 0.61 & 0.42 & 0.47 & 0.43 & 0.54 & 0.50 & 0.69 & 0.42 & 0.46 \\
R15          & 0.99 & 0.65 & 0.67 & 0.72 & 0.60 & 0.83 & 0.66 & 0.98 & 0.72 & 0.50 \\
S1           & 0.99 & 0.58 & 0.68 & 0.65 & 0.58 & 0.61 & 0.63 & 0.71 & 0.73 & 0.74 \\
S2           & 0.93 & 0.56 & 0.58 & 0.56 & 0.27 & 0.54 & 0.52 & 0.52 & 0.72 & 0.55 \\
Spiral       & 1.00 & 0.22 & 0.00 & 0.00 & 0.57 & 0.13 & 1.00 & 0.26 & 1.00 & 0.14 \\
Unbalance    & 1.00 & 0.98 & 1.00 & 1.00 & 0.78 & 0.97 & 0.99 & 0.97 & 0.85 & 0.90 \\
Vary density & 1.00 & 1.00 & 0.87 & 0.87 & 0.57 & 0.41 & 0.87 & 0.73 & 1.00 & 0.51 \\
\hline
Total Rank   & 25   & 111  & 100  & 97   & 133  & 120  & 144  & 120  & 113  & 132  \\
Avg rank     & 1.19 & 5.29 & 4.76 & 4.62 & 6.33 & 5.71 & 6.86 & 5.71 & 5.4  & 6.29 \\
\hline
\hline
{\color[HTML]{FE0000} Rank} &
  {\color[HTML]{FE0000} 1} &
  {\color[HTML]{FE0000} 4} &
  {\color[HTML]{FE0000} 3} &
  {\color[HTML]{FE0000} 2} &
  {\color[HTML]{FE0000} 9} &
  {\color[HTML]{FE0000} 6} &
  {\color[HTML]{FE0000} 10} &
  {\color[HTML]{FE0000} 6} &
  {\color[HTML]{FE0000} 5} &
  {\color[HTML]{FE0000} 8}
\end{tabular}%
}
\end{table}

\endgroup

\begingroup
\setlength{\tabcolsep}{3pt} 
\renewcommand{\arraystretch}{1.5} 
\centering

\begin{table}[H]
\caption{Comparison of the performance of DenMune with other nine algorithms, based on ARI-score, on fifteen real datasets.}
\label{Tab:ari-real}
\centering
\resizebox{\textwidth}{!}{%
\begin{tabular}{lcccccccccc}
\hline
{\color[HTML]{FE0000} Dataset} &
  {\color[HTML]{FE0000} DenMune} &
  {\color[HTML]{FE0000} NPIR} &
  {\color[HTML]{FE0000} CBKM} &
  {\color[HTML]{FE0000} RS} &
  {\color[HTML]{FE0000} FastDP} &
  {\color[HTML]{FE0000} FINCH} &
  {\color[HTML]{FE0000} HDBSCAN} &
  {\color[HTML]{FE0000} RCC} &
  {\color[HTML]{FE0000} Spectral} &
  {\color[HTML]{FE0000} KM++} \\
\hline
\hline
Appendicitis     & 0.56 & 0.55 & 0.19 & 0.19 & 0.52 & 0.25 & 0.42 & 0.20 & 0.00 & 0.23 \\
Arcene           & 0.09 & 0.10 & 0.10 & 0.10 & 0.10 & 0.01 & 0.10 & 0.03 & 0.10 & 0.10 \\
Breast cancer    & 0.88 & 0.88 & 0.55 & 0.67 & 0.87 & 0.06 & 0.86 & 0.12 & 0.00 & 0.86 \\
Optical digits   & 0.94 & 0.68 & 0.75 & 0.61 & 0.62 & 0.43 & 0.71 & 0.67 & 0.58 & 0.66 \\
Pendigits        & 0.83 & 0.48 & 0.63 & 0.56 & 0.30 & 0.58 & 0.61 & 0.46 & 0.50 & 0.60 \\
Ecoli            & 0.75 & 0.60 & 0.46 & 0.47 & 0.30 & 0.27 & 0.49 & 0.23 & 0.38 & 0.45 \\
Glass            & 0.26 & 0.23 & 0.25 & 0.25 & 0.22 & 0.31 & 0.17 & 0.20 & 0.30 & 0.13 \\
Iris             & 0.76 & 0.92 & 0.77 & 0.83 & 0.76 & 0.76 & 0.57 & 0.75 & 0.94 & 0.81 \\
MNIST            & 0.84 &  N/A$^{*}$  & 0.62 & 0.64 & 0.41 & 0.59 & 0.83 & 0.08 & 0.78 & 0.46 \\
Libras movement  & 0.32 & 0.25 & 0.31 & 0.26 & 0.26 & 0.26 & 0.23 & 0.23 & 0.27 & 0.22 \\
Robot navigation & 0.26 & 0.12 & 0.16 & 0.15 & 0.03 & 0.24 & 0.20 & 0.08 & 0.05 & 0.25 \\
SCC              & 0.66 & 0.62 & 0.54 & 0.68 & 0.45 & 0.51 & 0.48 & 0.61 & 0.56 & 0.47 \\
Seeds            & 0.69 & 0.75 & 0.74 & 0.71 & 0.48 & 0.56 & 0.58 & 0.71 & 0.42 & 0.54 \\
WDBC             & 0.49 & 0.66 & 0.41 & 0.53 & 0.60 & 0.46 & 0.41 & 0.20 & 0.00 & 0.46 \\
Yeast            & 0.13 & 0.11 & 0.16 & 0.14 & 0.17 & 0.13 & 0.15 & 0.18 & 0.10 & 0.11 \\
\hline
Total Rank       & 41   & 53   & 59   & 62   & 82   & 87   & 74   & 98   & 94   & 83   \\
Avg rank         & 2.93 & 3.79 & 4.21 & 4.43 & 5.86 & 6.21 & 5.29 & 7.00 & 6.7  & 5.93 \\
\hline
\hline
{\color[HTML]{FE0000} Rank} &
  {\color[HTML]{FE0000} 1} &
  {\color[HTML]{FE0000} 2} &
  {\color[HTML]{FE0000} 3} &
  {\color[HTML]{FE0000} 4} &
  {\color[HTML]{FE0000} 6} &
  {\color[HTML]{FE0000} 8} &
  {\color[HTML]{FE0000} 5} &
  {\color[HTML]{FE0000} 10} &
  {\color[HTML]{FE0000} 9} &
  {\color[HTML]{FE0000} 7}
\end{tabular}%
}
\end{table}
\endgroup

\begingroup
\noindent
\tiny
* NPIR failed to scale to adapt to MNIST dataset even on a cloud server with 128 GB memory.
\endgroup

\begin{landscape}

\begingroup
\setlength{\tabcolsep}{3pt} 
\renewcommand{\arraystretch}{2.0} 
\centering


\begin{table}[H]
\caption{Homogeneity \& Completeness scores of DenMune for the twenty-one synthetic datasets}
\label{Tab:denmune_homogeneity}
\centering
\resizebox{1.6\textwidth}{!}{%
\begin{tabular}{lccccccccccccccccccccc}
\hline
Dataset &
  A1 &
  A2 &
  Aggreg. &
  Compound &
  D31 &
  Dim32 &
  Dim128 &
  Dim512 &
  Flame &
  G2-10 &
  G2-30 &
  G2-50 &
  Jain &
  Mouse &
  Pathbased &
  R15 &
  S1 &
  S2 &
  Spiral &
  Unbalance &
  Varydensity \\
  \hline
  \hline
Homogeneity &
  0.97 &
  0.98 &
  0.99 &
  0.95 &
  0.96 &
  1 &
  1 &
  1 &
  1 &
  1 &
  0.94 &
  0.55 &
  1 &
  0.94 &
  0.94 &
  0.99 &
  0.99 &
  0.95 &
  1 &
  1 &
  1 \\
Completeness &
  0.99 &
  0.99 &
  0.99 &
  0.92 &
  0.96 &
  1 &
  1 &
  1 &
  1 &
  1 &
  0.94 &
  0.48 &
  1 &
  0.94 &
  0.84 &
  0.99 &
  0.99 &
  0.94 &
  1 &
  1 &
  1 
\end{tabular}%
}
\end{table}

\endgroup

\begingroup
\setlength{\tabcolsep}{3pt} 
\renewcommand{\arraystretch}{2.0} 
\centering
\begin{table}[H]
\caption{Homogeneity \& Completeness scores of DenMune for the fifteen real datasets}
\label{Tab:denmune_completeness}
\centering
\resizebox{1.6\textwidth}{!}{%
\begin{tabular}{lccccccccccccccc}
\hline
Dataset &
  Appendicitis &
  Arcene &
  Breast cancer &
  Optical digits &
  Pendigits &
  Ecoli &
  Glass &
  Iris &
  MNIST &
  Libras movement &
  Robot navigation &
  SCC &
  Seeds &
  WDBC &
  Yeast \\
  \hline
  \hline
Homogeneity &
  0.36 &
  0.1 &
  0.81 &
  0.96 &
  0.94 &
  0.67 &
  0.44 &
  0.8 &
  0.9 &
  0.78 &
  0.52 &
  0.8 &
  0.69 &
  0.42 &
  0.34 \\
Completeness &
  0.39 &
  0.06 &
  0.8 &
  0.92 &
  0.82 &
  0.75 &
  0.35 &
  0.82 &
  0.82 &
  0.59 &
  0.37 &
  0.84 &
  0.7 &
  0.51 &
  0.23
\end{tabular}%
}
\end{table}
\endgroup
\end{landscape}

\bibliography{article}{}
\bibliographystyle{elsarticle-num}

\end{document}